\documentclass[10pt,journal,compsoc]{IEEEtran}

\usepackage{textcomp}
\usepackage{stfloats}
\usepackage{url}
\usepackage{verbatim}
\usepackage{graphicx}
\usepackage{cite}
\usepackage{ragged2e}
\usepackage{color}
\usepackage{booktabs}
\usepackage{multirow}
\usepackage{hyperref}
\usepackage{amssymb}
\usepackage{amsmath}
\usepackage{bbding}

\usepackage{color,array}

\usepackage{graphicx}

\usepackage{xcolor}
\usepackage[acronym, nonumberlist]{glossaries}
\definecolor{blueacro}{rgb}{0.01, 0.3 0.6}
\definecolor{greencite}{rgb}{0.01, 0.5 0.3}
\hypersetup{
	colorlinks,
	linkcolor={blueacro},
	citecolor={greencite},
	urlcolor={blue!80!black}
}
\usepackage{colortbl}

\usepackage{hhline}
\usepackage{subfiles}

\usepackage{comment}

\usepackage{comment}

\usepackage{import}

\usepackage{enumerate}
\usepackage{booktabs}

\usepackage{float}
\usepackage{subcaption}
\usepackage[format=plain,
labelfont=it,
textfont=it]{caption}

\usepackage{amsmath,amsfonts, bm} 
\usepackage{graphicx} 
\usepackage{multirow} 
\usepackage{rotating} 
\usepackage{url} 

\usepackage[linesnumbered,ruled,vlined]{algorithm2e} 
\usepackage{amsmath}  
\usepackage{amssymb}  
\usepackage{bm}       
\usepackage{mathtools} 

\DeclareMathOperator{\sign}{sign}
\DeclareMathOperator{\abs}{abs}
\DeclareMathOperator{\vecc}{vec}

\usepackage{titlesec}
\usepackage{microtype}
\usepackage{makecell}

\usepackage{pifont}
\newcommand{\cmark}{\ding{51}}%
\newcommand{\xmark}{\ding{55}}%

\newcommand{\underscore}{\ifmmode\_\else\textunderscore\fi}

\newcommand{\idf}[1]{#1}

\def \t {{^{t}}}

\def\incrmodel{{v}}
\def\incrmodelparams{{\theta}}
\newcommand{\Tau}{\mathcal{T}}

\def\Xc{{\mathcal{X}}}
\def\Yc{{\mathcal{Y}}}
\def\Zc{{\mathcal{Z}}}
\def\Dc{{\mathcal{D}}}
\def\Lc{{\mathcal{L}}}
\def\Nc{{\mathcal{N}}}

\newcommand{\Fc}{\mathbf{F}}

\def\Ab{{\mathbf{A}}}
\def\Bb{{\mathbf{B}}}
\def\Fb{{\mathbf{F}}}

\def\Mb{{\mathbf{M}}}

\def\Mb{{\mathbf{M}}}
\def\Wb{{\mathbf{W}}}

\def\x{{\mathbf{x}}}
\def\y{{\mathbf{y}}}
\def\z{{\mathbf{z}}}

\def\Rb{\mathbb{R}}
\def\Eb{\mathbb{E}}
\def\Uc{\mathcal{U}}

\def\loradone{{a}}
\def\loradtwo{{b}}
\def\lorarank{{r}}

\newcommand{\circshift}{\operatorname{rshift}}

\newcommand{\mypowsymbol}{\boldsymbol{\delta}}
\newcommand{\myfreqsymbol}{\boldsymbol{\omega}}

\newcommand{\mypow}{\operatorname{pow}}
\newcommand{\mypowsigned}{\operatorname{pow-s}}
\newcommand{\mypowsignedlearned}{\operatorname{pow, \mypowsymbol}}

\newcommand{\mycos}{\operatorname{cos}}
\newcommand{\mycosfreq}{\operatorname{cos, \myfreqsymbol}}

\newcommand{\paramclassif}{\phi}
\newcommand{\alphalora}{\alpha}

\usepackage{enumerate}

\makeatletter
\def\endthebibliography{%
	\def\@noitemerr{\@latex@warning{Empty `thebibliography' environment}}%
	\endlist
}
\makeatother

\begin{document}
	
	\title{Deep Generative Continual Learning using Functional LoRA: FunLoRA}
	
	\author{Victor Enescu, Hichem Sahbi
		\IEEEcompsocitemizethanks{\IEEEcompsocthanksitem V. Enescu and H. Sahbi are with Sorbonne University, CNRS, LIP6, F-75005 Paris, France. 
			Corresponding authors:  victor.enescu@lip6.fr, hichem.sahbi@lip6.fr}
	}

	\IEEEtitleabstractindextext{
		\begin{abstract}
			\justifying
		Continual adaptation of deep generative models holds tremendous potential and critical importance, given their rapid and expanding usage in text and vision based applications. Incremental training, however, remains highly challenging due to catastrophic forgetting phenomenon, which makes it difficult for neural networks to effectively incorporate new knowledge. A common strategy consists in retraining the generative model on its own synthetic data in order to mitigate forgetting. Yet, such an approach faces two major limitations: (i) the continually increasing training time eventually becomes intractable, and (ii) reliance on synthetic data inevitably leads to long-term performance degradation, since synthetic samples lack the richness of real training data. In this paper, we attenuate these issues by designing a novel and more expressive conditioning mechanism for generative models based on low rank adaptation (LoRA), that exclusively employs rank 1 matrices, whose reparametrized matrix rank is functionally increased using carefully selected functions — and dubbed functional LoRA: FunLoRA. Using this dynamic conditioning, the generative model is guaranteed to avoid catastrophic forgetting and needs only to be trained on data from the current task. Extensive experiments using flow-matching based models trained from scratch, showcase that our proposed parameter-efficient fine-tuning (PEFT) method surpasses prior state-of-the-art results based on diffusion models, reaching higher classification accuracy scores, while only requiring a fraction of the memory cost and sampling time.
 		\end{abstract}
		
\begin{IEEEkeywords}
	Continual Learning, Generative Models, Flow-Matching, Rehearsal Learning, Low-Rank Adaptation
\end{IEEEkeywords}}

	\maketitle
	\section{Introduction}\label{sec:intro}
	\IEEEPARstart{O}{ne} of the primary objectives of deep learning, consists in developing smart agents, capable of continuously learning and interacting with their surrounding environments — thereby mimicking the adaptive capabilities of humans. Despite already exceeding our performance on various tasks, the continuous acquisition of new concepts in neural networks leads to a severe degradation in performance, a phenomenon commonly referred to as catastrophic forgetting~\cite{mccloskey_catastrophic_forgetting}. More specifically, the optimization of weights for a current task inadvertently overwrites representations learned from previous data distributions thereby causing forgetting — a challenge that is further exacerbated by sequential task learning.\\
	\indent Initial approaches to solve this sequential paradigm known as Continual Learning (CL)~\cite{thrun1995lifelong, ring1997child, parisi2019continual}, primarily focused on discriminative models, due to their involvement in the core decision-making process of autonomous agents and their shorter training time.  However, the unprecedented rise of deep generative models \cite{chatgpt3, stable_diffusion, llama} has marked a new era in deep learning, creating the need for novel strategies to adapt these increasingly indispensable networks. Seminal approaches such as low rank adaptation (LoRA) \cite{lora}, Adapters \cite{adapters} and Prompt Tuning \cite{prompt_tuning} introduced solutions known as Parameter-Efficient Fine-Tuning (PEFT)~\cite{peft_survey}, initially used to adapt transformer~\cite{attention} based large language models. Nevertheless those methods can readily be extended to visual models~\cite{peft_conv, vpt, lora_clr}, and warrant further research as continual learning trends and requirements evolve. Indeed, early continual learning strategies largely relied on storing and replaying exemplars from past tasks~\cite{icarl, dytox}, so as to retrain discriminative networks~\cite{lecun_cnn, vit}. However these solutions are often constrained by memory limitations~\cite{online_continual_learing_embedded, tiny_ml_continual_learning} and raise privacy concerns~\cite{privacy_cl_1, privacy_cl_2}. As such, a rise in popularity has been observed for exemplar-free class-incremental learning (EFCIL) methods~\cite{fetril, plastil, learning_without_forgetting}, that explicitly do not store any data from previous tasks in memory. Among these approaches, generative models are particularly appealing, as they can — in principle — produce an infinite amount of data via sampling. Nevertheless, the generative model itself is also prone to catastrophic forgetting, and must also therefore be trained incrementally. \\ 
	\indent Early attempts to solve this challenge employ regularization techniques on the generative model~\cite{variational_continual_learning, continual_learning_in_gans, continual_learning_diffusion_distillation}, so as to maintain image consistency through tasks. However they face the plasticity-rigidity dilemma, leading to substantial forgetting or a severe inability to learn new classes. Other solutions  employ self-replay strategies~\cite{mergan, pseudo_rehearsal, ddgr, gppdm, guide_diffusion, jdcl, continual_learning_diffusion_distillation}, which involve re-training the generative model on its own conditional synthetic data — so as to mitigate forgetting. However, this solution not only results in a linearly increasing training time for the generative model — since it must revisit its own synthetic data — but also a severe and unavoidable performance degradation as observed in~\cite{forgetting_flowers, on_the_stability_of_iterative_retraining, self-consuming_generative_models_go_mad}. Alternative solutions consider starting from pre-trained foundation models, such as stable-diffusion~\cite{class_incremental_diffusion_pretraind_gaussian, diffusion_og, stable_diffusion}, that can be conditioned using the  prompting mechanism~\cite{class_incremental_learning_diffusion_pretrained, class_incremental_diffusion_pretraind_gaussian, reality_check_pretraining_popescu}, or by employing PEFT methods~\cite{c_lora_diffusion, lora_continual_personalization_diffusion, flexible_customization_dm, lora_continual_dm_deja} such as low rank adaptation~\cite{lora}. Nonetheless, these foundation models demand substantial compute and training time, which may not meet continual learning constraints. As such, incrementally training generative models from scratch is a standing and underexplored challenge in continual learning. An ideal solution would only allocate a few more extra conditional parameters per class while exclusively training on a new task — as this would ensure the parameter cost remains minimal and there is no forgetting, since the architecture is dynamic.\\
	\indent In order to address these aforementioned issues, we present a novel PEFT method for continual adaptation of conditional generative models. The core of this contribution relies in a LoRA reparametrization exclusively using rank 1 matrices, so as to remain very frugal in terms of memory footprint. Nonetheless to ensure those (LoRA) matrices are expressive, their rank is functionally increased (with an equivalent number of parameters) using carefully selected functions, most notably trigonometric and exponential functions. As such, this method can be seen as a \emph{"functional"} LoRA (dubbed FunLoRA), that greatly increases expressiveness using rank 1 matrices only. Furthermore, we do not focus on the attention layer when reparametrizing with LoRA, but on the convolutional layers instead — as they simultaneously incorporate the class-conditional and time information in the employed U-Net~\cite{unet} backbone (unlike attention layers). Indeed, as observed in the experiments, LoRA applied to convolutional layers achieves better results, especially in incremental learning scenarios where there is very little data available to train the attention layers, which cannot learn generalizable patterns across different tasks.  Based on this novel PEFT method, we also locate the most important layers to adapt in the U-Net backbone — drastically reducing the parameter cost required for learning new classes. Finally, we specifically focus on incrementally conditioning flow-matching models~\cite{flow_matching, rectified_flow, stochastic_interpolant, cfm_tong}, which are faster to sample than diffusion models~\cite{diffusion_og, ddpm, ddim}, while exhibiting equivalent image quality. Methods closest to our work adapt diffusion models incrementally using self-replay and regularization strategies~\cite{guide_diffusion, jdcl, ddgr, hal_continual_diffusion, class_prototype_dm}, and we demonstrate stronger performance on widely used benchmarks such as CIFAR and ImageNet, achieving superior classification accuracy using only a fraction of their parameter cost. \\
	\indent Considering all the aforementioned issues, the main contributions of this paper include: 
	\begin{enumerate}
		\item A novel LoRA based PEFT method optimized for adapting generative models in continual learning. This method — dubbed FunLoRA — allocates a minimal number of parameters through LoRA reparametrizations, and then functionally increases the rank of the resulting matrix for increased expressiveness. 
	
		\item \textcolor{black}{An innovative flow-matching based solution for generative continual learning, that is very fast to adapt and does not suffer from any forgetting. This proposed method does not require the generative model to retrain on its own synthetic data, and reaches performances very close to a generative upper bound trained in a multi-task setting — while only allocating up 2.3\% extra parameters.}
		
		\item A new approach to identify most important layers when training generative models incrementally, which greatly reduces memory footprint and training time in continual learning scenarios. Most notably, we focus on the convolutional layers of the standard U-Net backbone since it incorporates both class and time information.
		
		\item  Extensive experiments through different benchmarks showing the outperformance of our proposed incremental generative method against state-of-the-art related works — while using only a fraction of their memory cost. Most notably, on the CIFAR100~\cite{cifar} benchmark, our method can even outperform pre-trained stable diffusion models in continual learning, using orders of magnitude less parameters and training data.
	\end{enumerate}

	\section{Related Work}
	\subsection{Continual Learning.} Three different categories of continual learning methods exist in the literature: \emph{(1) dynamic architectures} \cite{dynamic_cl_1, hat, den, pnn} which allocate different subnetwork parameters for different tasks, \emph{(2) regularization-based} methods \cite{continual_learning_through_synaptic_inelligence, ewc,memory_aware_synapses, reg_cl_4} that balance plasticity and rigidity in order to prevent significant updates of important parameters from previous tasks and thereby avoid catastrophic forgetting; this category also includes optimization based  methods that project the parameters of new tasks in orthogonal null spaces \cite{adam_nscl, omw_orthogonal_gradient, orthogonal_cl_3, orthogonal_cl_4} in order to avoid catastrophic interference. The third category of methods includes \emph{(3) replay} which either directly stores training data from previous tasks in a memory buffer~\cite{online_continual_learning, mem_replay_cl_2, forgetting_metric, mem_replay_cl_4} or train a generative model in order to synthesize data from the same distribution as the actual training data~\cite{gan_cl_1, variational_continual_learning}. 
	This category in particular, has attracted a lot of attention in the literature with various generative models being applied to incremental learning including GANs, \cite{gan_cl_1, dgr, gan}, VAEs \cite{variational_continual_learning, vae_cl_2, vae}, normalizing flows (NFs) \cite{task_agnostic_cl, pseudo_rehearsal, nf_kobyzev}, and diffusion models (DMs)~\cite{class_prototype_dm, ddgr, diffusion_og, hal_continual_diffusion}. Nonetheless, these incrementally learned generative  models are known to be computationally demanding with a significant increase in time and memory footprint as the number of tasks increases. Indeed, early incremental generative models based on GANs and VAEs, use regularization techniques \cite{mergan, variational_continual_learning, continual_learning_in_gans, continual_learning_diffusion_distillation}, in order to ensure that the image generated from a fixed latent noise vector, remains (almost) identical in the new generative model. However, in practice, the incremental generative model is confronted to the plasticity-rigidity dilemma, and substantial forgetting occurs when the model attempts to learn new tasks. Similarly, this forgetting is also unavoidable in self-replay strategies, which retrain the generative model on its own conditional synthetic data \cite{mergan, pseudo_rehearsal, ddgr, gppdm, guide_diffusion, jdcl, continual_learning_diffusion_distillation}. Indeed, this repeated self retraining is known to cause highly reduced diversity or complete degradation \cite{forgetting_flowers, on_the_stability_of_iterative_retraining, self-consuming_generative_models_go_mad} of the generative capabilities, and we will avoid this approach in our contribution. Alternatively, to reduce training time and memory cost, the generative model can be trained in the feature space \cite{generative_feature_replay_gan} VAE \cite{cgil, brain_inspired_replay_vae, closed_loop_transcription_incremental_vae}, NFs \cite{task_agnostic_continual_learning_nf} to DM \cite{dm_replay_features} of a classifier. However, the resulting models can no longer synthesize actual training data (images), and often those solutions employ class/task dependent generative models~\cite{cgil, dm_replay_features}, which does not pose any challenge from the point of view of continual learning.
	
	\indent On the other hand, starting from very large pre-trained generative models such as stable diffusion can bypass the requirement of training the models on new tasks~\cite{class_incremental_learning_diffusion_pretrained, reality_check_pretraining_popescu}, by generating images based on prompt conditioning. Most notably \cite{class_incremental_diffusion_pretraind_gaussian} estimates class prototypes in the latent space of a pre-trained diffusion model to filter them based on a threshold. However, to reach best performances, it still relies on Parameter-Efficient Fine-Tuning (PEFT) techniques, which are further discussed in the following section. 
	
	\subsection{PEFT for Continual Learning.} Numerous PEFT techniques have recently emerged as powerful solutions to continual learning, as they require only very little parameters, and do not suffer from overfitting when applied to large pre-trained models --- a phenomenon severely exacerbated at low data regimes commonly encountered in incremental learning. Furthermore, relatively faster training times can be achieved since gradients are only computed for a portion of the models parameters, which is an important aspect of continual learning, and contributes towards tractable real time, and memory efficient learning. This is especially true for prompt-learning methods \cite{coop, cocoop, coda_prompt, dual_prompt, attri_clip, cgil, star_prompt, vpt_continual, eclipse_visual_continual} commonly applied to visual language models such as CLIP \cite{clip}, which constitute a very interesting approach, as they only modify prompts given as inputs to transformer networks. More specifically, extra prompting parameters fed to the text encoder are optimized to improve the classification accuracy, by maximizing their cosine similarity with image representations extracted by the vision encoder. 	In order to mitigate interference and forgetting when learning those incremental prompts, orthogonality constraints can be employed as in \cite{coda_prompt, dual_prompt, attri_clip, vpt_continual}. Alternatively, instead of preventing significant changes to the prompting parameters, generative/feature replay can be performed \cite{cgil, star_prompt, prompt_feature_replay_continual},  in which case the prompting parameters can be retrained without any constraint. More specifically, \cite{cgil, cgil_cond} utilize variational autoencoders for memory replay to retrain prompting parameters, and \cite{star_prompt, prompt_feature_replay_continual} stores gaussian mixture models to resynthesize artificial features. Similarly, \cite{slca, slca++} leverage feature based replay on gaussians, but combines it with adapters modules introduced in attention layers instead of prompts. Alternatively, numerous LoRA based methods also target attention layers \cite{pilora, sd_lora, o_lora, c_lora, lori, critical_param_lora, fm_lora, dc_lora, orthogonal_lora_lie_cl, lora_critical_parameter_change_cl, tree_lora_cl, inflora} for continual learning, but mostly focus on pre-trained ViT models for continual image classification \cite{fm_lora, c_lora, pilora, lora_critical_parameter_change_cl, tree_lora_cl}.

 On the other hand, solutions not specifically tailored for continual classification adapt pre-trained stable diffusion models with LoRA~\cite{c_lora_diffusion, lora_continual_personalization_diffusion, flexible_customization_dm, lora_continual_dm_deja} for image customization.  Prior to these, GAN based methods were extensively studied for style transfer and few shot image generation, but by starting from smaller pre-trained models \cite{gan_memory, lfs_gan, kernel_modulation, adam_few_shot}. Amongst them, GAN-memory \cite{gan_memory} — a precursor to Parameter-Efficient Fine-Tuning — applies feature modulation on convolutional layers, and achieves good performance in continual classification. In this work, we also focus on adapting convolutional layers on generative models, but by introducing a modern LoRA approach that functionally increases the rank of the matrices associated to each layer in the network. As such, our incrementally trained generative model does not suffer from forgetting, and achieves far superior performance compared to closest works using diffusion models for continual learning~\cite{dgr, ddgr, guide_diffusion, jdcl}, while requiring less parameters. Furthermore, our method leverages a modern flow-matching paradigm, better suited than diffusion for continual classification, as it achieves faster sampling for equivalent image quality. Most notably, our proposed solution can also outperform solutions leveraging pre-trained stable diffusion models~\cite{class_incremental_diffusion_pretraind_gaussian}, which are notoriously slower to sample, and are trained on orders of magnitude more data.

	\section{Background}
	
	\def\loradone{{a}}
	\def\loradtwo{{b}}
	\def\lorarank{{r}}
	\def\kernelsize{{s}}

	\subsection{Conditional Flow Matching}
	Conditional Flow matching~\cite{flow_matching, rectified_flow, stochastic_interpolant, cfm_tong}  is a generative paradigm inspired from continuous normalizing flows \cite{node_nf}, that is simulation free and very fast to train, as it does not require solving ordinary differential equations in the forward pass. It consists in learning a vector field $v : [0, 1] \times \Rb^{d} \rightarrow \Rb^{d}$, that is used to construct a time-dependent diffeomorphic mapping called a flow $\phi: [0, 1] \times \Rb^{d} \rightarrow \Rb^{d}$ between an ambient space $\Xc$ and a latent space $\Zc$.  More generally, the neural network $v_{\theta}(t, \x)$ parametrized by $\theta$, learns the conditional vector field $u_{t}(\x | \z)$ by minimizing a mean squared error as shown in Eq~\ref{eq:cfm_eq}~\cite{flow_matching, cfm_tong}.
	\begin{equation}
		\Lc_{CFM} = \Eb_{t \sim \Uc({\bf{0}, \bf{I}}), q(\z), p_{t}(\x|\z)} 
		\lVert
		v_{\theta}(t, \x) - u_{t}(\x | \z)
		\rVert_{2}^{2}
		\label{eq:cfm_eq}
	\end{equation}
	where $q(\z)$ is a gaussian prior $\Nc(\bf{0}, \bf{I})$, and  $p_{t}(\x|\z) $ is \textbf{known probability path}  that is chosen and defined for each sample in the dataset. In general, the probability path between an image $\x_{0}$, and a random noise $\z$ is defined as:
	\begin{equation}
		\x = (1 - t) \x_{0} + t \z 
		\label{eq:ot_displacement_map}
	\end{equation}
	
	\noindent
	where $\x$ is the noisy data sample used in Eq.~\ref{eq:cfm_eq}. It can be noted that this formulation shares similarities with the \emph{"paths"} used in diffusion models, but in this case they are much simpler and straighter, which means they are faster to integrate when sampling than in the case of diffusion models. Indeed,  Eq.~\ref{eq:ot_displacement_map} corresponds to the optimal transport (OT) displacement map \cite{ot_map}, and this property is very interesting for continual learning where faster sampling is of utmost importance. In the following sections, we will simply refer to conditional flow matching as \emph{"flow matching"}, and the word \emph{"conditional"} will indicate the conditioning on class labels $\y$. \\

	\subsubsection{LoRA} \label{sec:lora_reminder_with_conv}
	Given the pre-trained weight $\Wb_{0} \in \Rb^{\loradone \times \loradtwo}$ of a fully connected layer, LoRA seeks to learn an extra weight matrix  $\Delta \Wb \in \Rb^{\loradone \times \loradtwo}$ that will modify the initial values of $\Wb_{0}$ for higher expressiveness, and that is reparametrized as the product of two low rank matrices $\Ab \in \Rb^{\loradone \times \lorarank}$ and $\Bb \in \Rb^{\lorarank \times \loradtwo}$, so that $\Delta \Wb = \Ab \Bb$. This reparametrization is advantageous since the number of parameters in $\Delta \Wb \in \Rb^{\loradone \times \loradtwo}$ is much smaller than in $\Wb_{0}$. At inference, an input $\x$ will interact with both weight matrices so that:
	\begin{equation}
		\x'  = (\Wb_{0} + \Delta \Wb)\x =  \Wb_{0}\x + \Ab\Bb\x
		\label{eq:lora}
	\end{equation}
	where the rank $r$ is a hyperparameter that must be tuned, and can be seen as the lower dimension in which the data is down-projected ($\Bb\x$), before it is \textcolor{black}{reconstructed into} its initial dimension $(\Ab \cdot (\Bb\x))$. Usually, LoRA is applied to fully connected layers in attention blocks, however it can also be extended to convolutional layers with weight parameters denoted as $\Wb_{0} \in \mathbb{R}^{C_{out} \times C_{in} \times \kernelsize \times \kernelsize}$,  where $C_{out}$ and $C_{in}$ respectively denote the number of output and input feature channels, and $ \kernelsize$ the kernel size. Most notably,  by defining the parameters $a_{conv}, b_{conv}, r_{conv}$ as the convolutional equivalents of $a, b, c$ where $a_{conv} = C_{out}$ and $b_{conv} = C_{in} \cdot \kernelsize \cdot \kernelsize$, the matrices  from Eq~\ref{eq:lora} can be rewritten as $\Ab \in \Rb^{a_{conv} \times \lorarank_{conv}}$ and $\Bb \in \Rb^{ \lorarank_{conv} \times b_{conv}}$, and their product is reshaped to match the dimension of the initial weight tensor $\Wb_{0}$. Furthermore, the variable $r_{conv}$ can be customized, and is often set to $\lorarank_{conv} = \lorarank \cdot \kernelsize$ so as to add a dependency on the kernel size in the bottleneck dimension. In general, the matrix $\Bb$ is initialized to zero, and the matrix $\Ab$ follows a Kaiming normal initialization. The zero initialization ensures the identity function is initially preserved since the product $\Ab \Bb  = \mathbf{0}$, and the \textcolor{black}{Kaiming initialization ensures that non-zero gradients can flow into both matrices during optimization.} \\

	\section{Proposed Method} \label{sec:method}
	\subsection{Problem Formulation.}
	We consider a continual learning scenario, where a conditional flow matching model $\incrmodel_{\incrmodelparams, \y}(.)$ parametrized by $\incrmodelparams$ and conditioned on labels $\y$ is trained (using Eq.~\ref{eq:cfm_eq}) on a stream of data divided into $T$ distinct groups called tasks. Each task $\Tau\t$ can be seen as a dataset $\Dc\t$ = $\{(\x^{t}_{i}, \y^{t}_{i})\}_{i}$ made of image-label pairs,  with several distinct classes constituting a task. We denote $\Yc^{t}$ as the label space of task $t$,  verifying the condition $\Yc\t \cap \Yc^{t'} = \emptyset $ for  $t \neq t'$, meaning there are no overlapping classes between tasks. Additionally, we denote $\Yc^{1:t} = \cup_{t'=1}^{t} \Yc^{t'}$ as the union of all label spaces up to task $t$. Following this notation, incremental learning consists in updating the parameters $\incrmodelparams$ on a current task $t$ while minimizing catastrophic forgetting on all the previous classes in $\Yc^{1:t-1}$.  
	Furthermore, to assess the performance of this incremental generative model, we consider a classifier network $c_{\paramclassif}(.)$ parametrized by $\paramclassif$, more specifically its accuracy on a test set with labels including all classes $\in \Yc^{1:t}$. This setup, where the classifier network $c_{\paramclassif}(.)$ is trained without any constraint on fully synthetic data, is possible thanks to the proposed incremental generative model.

	\subsection{LoRA based approach} \label{sec:lora_based_approach}
	Concerning the backbone of the generative model, which is a standard convolutional U-Net popularized with diffusion models \textcolor{black}{—} we only focus on adapting the convolutional layers in subsequent tasks \textcolor{black}{—} since the conditioning and timestep information is only applied on these portions of the network.
	Most notably, given the weight parameters $\Wb_{0} \in \mathbb{R}^{C_{out} \times C_{in} \times \kernelsize \times \kernelsize}$  of a convolutional layer, we propose to perform a LoRA decomposition to adapt the magnitude of each filter  using two matrices $\Ab \in \mathbb{R}^{C_{out} \times 1}$ and $\Bb \in \mathbb{R}^{1 \times C_{in}}$ so that:
	\begin{equation}
		\Fb = \Ab  \Bb
	\end{equation}
	is used to modify the weight kernels from $\Wb_{0}$ as
	\begin{equation}
		\Wb = \Wb_{0} \odot \Fb
	\end{equation}
	where $\Fb$ is reshaped to have dimensions $\in \Rb^{C_{out} \times C_{in}}$. This strategy of multiplying the convolutional filters with a matrix $\Fb$ (without using a LoRA based on $\Ab$ and $\Bb$) is inspired from \cite{gan_memory}, and is called feature modulation, and was historically used for style transfer (prior to the advent of stable diffusion and LoRA). 
	
	Afterwards, the matrix $\Wb$ which had its kernel rescaled is used when performing a convolution with the input data. Furthermore, since we focus on incremental learning (which benefits from reduced memory cost), the  shared dimension between $\Ab$ and $\Bb$ (\emph{i.e} the rank) is chosen to be 1; these matrices are initialized with ones to ensure the identity operation is preserved during the first iteration. Indeed, as will be shown in the following sections, this also makes it possible to directly calculate the importance of each layer in the neural network, and can be more expressive than standard addition when using rank 1 matrices. Furthermore, in order to further facilitate class-incremental learning of our generative model, we also condition each matrix $\Ab$ and $\Bb$ on a label $\y$ so that 
	\begin{equation}
		\Fb_{\y} = \Ab_{\y}  \Bb_{\y}
		\label{eq:feature_modulation_vanilla_decomp}
	\end{equation}
	and consequently 
	\begin{equation}
		\Wb_{\y} = \Wb_{0} \odot \Fb_{\y}
		\label{eq:feature_modulation_vanilla_decomp_product}
	\end{equation}
	which is the equivalent of learning a different sub-instance of a neural network for each class. This formulation has the advantage of dissociating the learning of classes from tasks, and is the equivalent to learning a single class per task, which is the most challenging scenario in continual learning. Furthermore, it allows for efficient parameter sharing/reuse if the distribution of a certain class $\y$ changes as only the parameters associated to the current class would be duplicated / fine-tuned. \\
	
	\subsection{Functional LoRA}
	\noindent  In order to further increase the expressiveness of the proposed method, we consider applying several functions $\{f_i\}_i$ to the matrices $\Ab_{\y}$ and $\Bb_{\y}$ (that have a minimal rank of 1), before or after taking their product $\Ab_{\y} \Bb_{\y}$. \textcolor{black}{This increases the rank} of a final \emph{"functional"} matrix $\Fc_{\y} \in \Rb^{C_{out} \times C_{in}}$ that will be used in the LoRA reparametrization, and defined as: 
	\begin{equation}
		\Fc_{\y} = \frac{1}{p}\sum_{i=1}^{p}  \alpha_{i} f_{i}(\Ab_{\y}, \Bb_{\y})
		\label{eq:fn_sum_general}
	\end{equation}
	where $p$ is the total number of distinct functions $\{f_i\}_i$ applied, and $\alpha_{i}$ is a learnable ponderation factor associated to each function $f_{i}$. As a result  the rank of the matrix $\Fc_{\y}$ increases according to the value of $p$, but also depending on the choice of the functions $\{f_i\}_i$.  Different possibilities are available for those functions, and may exhibit different advantages as will be explained in the following sections. Those functions may include circular shifts, element-wise powers, or trigonometric functions. \\

	\noindent \textbf{Circular shifts.}  A right circular shift of a vector $\Mb$ by $i$ elements, denoted $	\circshift(\mathbf{M}, i)$, is defined using an $m \times m$ \textbf{permutation matrix} $P_R^{(i)}$. This matrix multiplication rearranges $\mathbf{M}$'s elements: the last $i$ elements move to the front, preserving their order, while the first $m-i$ elements shift to the end. The operation is simply $	\circshift(\Mb, i) = P_R^{(i)} \Mb$. The matrix $P_R^{(i)}$ is specifically crafted to select the final $i$ elements for the initial positions and shift the remaining elements accordingly. Naturally, this function is trivially extended to the transposed matrix $\Mb^{\top} \in \Rb^{m \times 1}$ in which case the transpose function is implicitly applied before and after applying the $\circshift(., i)$  function. Then, we can define a corresponding function $f^{\circshift}_{i}$ that applies this circular shift to two matrices $\Ab \in \Rb^{C_{out} \times 1}$ and $\Bb \in \Rb^{1 \times C_{in}}$ (used in LoRA) before multiplying them so that:
	\begin{equation}
		f^{\circshift}_{i}(\Ab, \Bb) = \circshift(\Ab, i) \times  \circshift(\Bb, i)
	\end{equation}
	outputs a matrix with dimension in $ \Rb^{C_{out} \times C_{in}} $. By successively applying this function with different values for the circular shifts \textcolor{black}{— one may show} that it is possible to increase the rank of the final functional matrix $\Fc_{\y, p}^{\circshift}$ defined as:
	\begin{equation}
		\Fc_{\y, p}^{\circshift} = \frac{1}{p}\sum_{i=1}^{p}  \alpha_{i} f_{i}^{\circshift}(\Ab_{\y}, \Bb_{\y})
	\end{equation}
	in which case the class conditional matrices $\Ab_{\y}$ and $\Bb_{\y}$ are employed for the LoRA parametrization. Theoretically speaking (and as will be seen in the experiments), assuming $C_{out}$, $C_{in} > p$,  the rank of final matrix $\Fc_{\y, p}^{\circshift}$ is generally equal to $p$, which confirms that this functional parametrization successfully increases the rank of the subsequent matrix used in LoRA. Nonetheless, as will be confirmed through experiments, the expressiveness is not the highest, and other functional parametrizations described in the following sections may exhibit superior performance. \\

	\begin{figure*}[!htb]
		\includegraphics[width=0.90\textwidth]{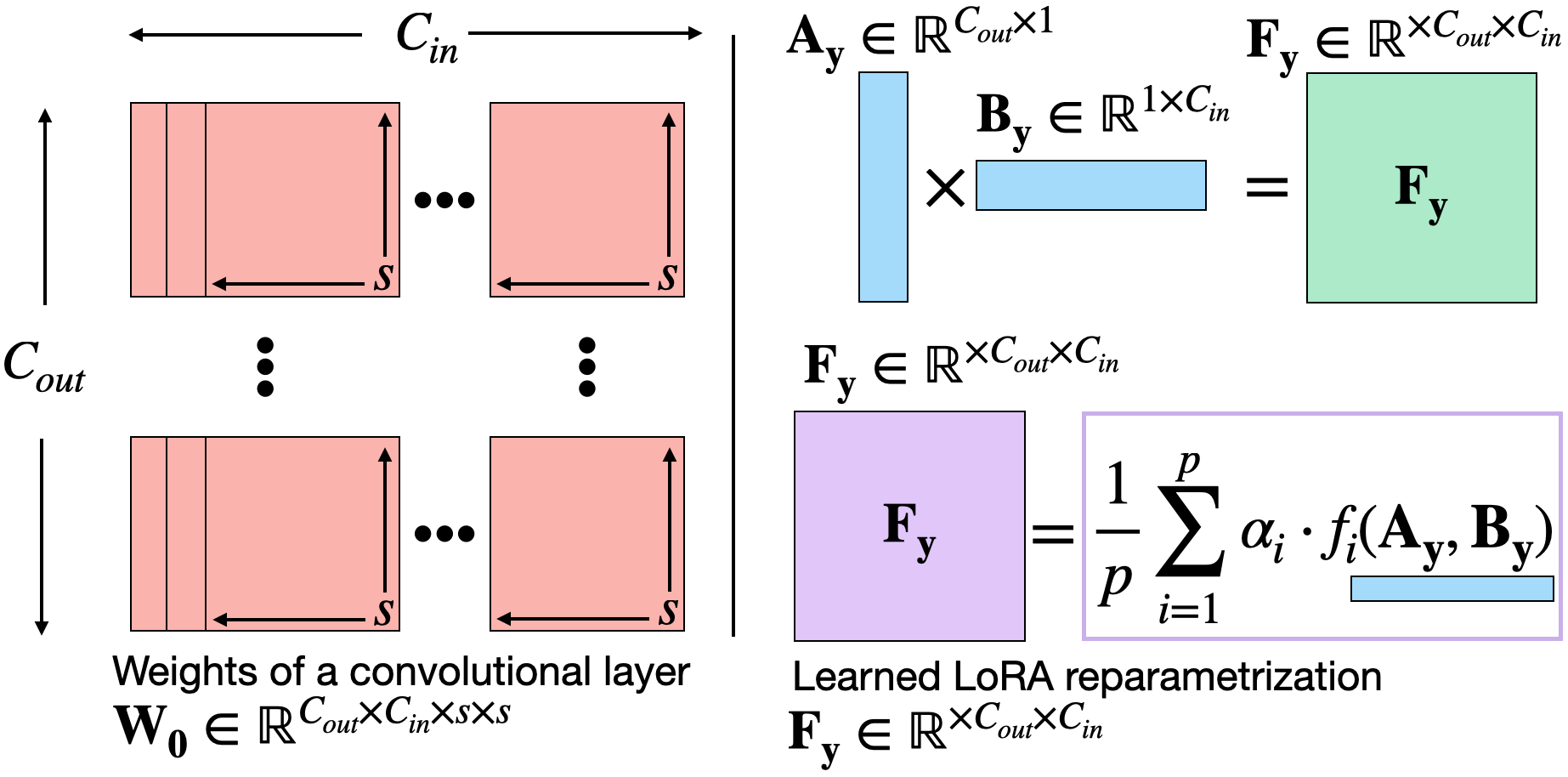}
		\centering
		\caption{Visualization of the proposed LoRA reparametrization applied to rescale all the filters in a convolutional layer. Given the convolutional weights $\Wb_{0} \in \mathbb{R}^{C_{out} \times C_{in} \times \kernelsize \times \kernelsize}$, we learn two low rank matrices $\Ab_{\y} \in \Rb^{C_{out} \times 1}$ and $\Bb_{\y} \in \Rb^{1 \times C_{in}}$ for each class so that their product gives a matrix $\Fb_{\y} \in \Rb^{C_{out} \times C_{in}}$ which spans all filters, and can be used to modify their values. To further increase the rank of this matrix $\Fb_{\y}$ (which is one by default), functions $\{f_i\}_i$ are applied to each entry in $\Fb_{\y}$ or to $\Ab_{\y}$ and $\Bb_{\y}$ separately, before taking the product.}
		\label{fig:lora_feature_modulation_conv}
	\end{figure*}
	\begin{figure*}[!htb]
		\includegraphics[width=0.75\textwidth]{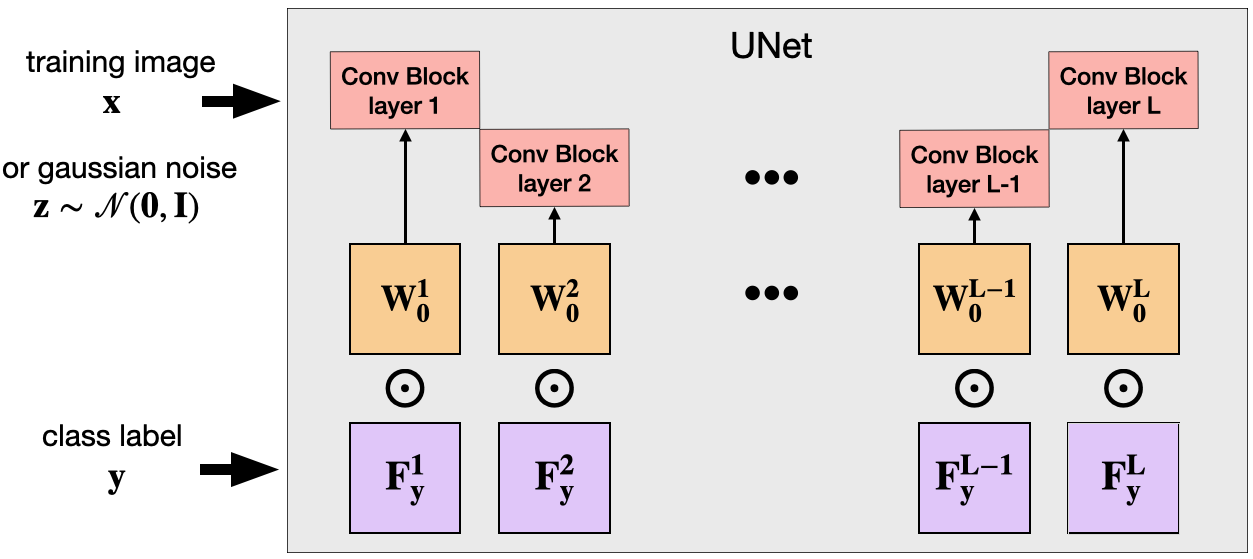}
		\centering
		\caption{Illustration of the U-Net whose convolutional layers with index $l$, are adapted using the class conditional functional matrix $\Fc^{l}_{\y}$. Exceptionally in this Figure and to avoid cluttered notations, layer indices ($l$) are used as superscripts.}
		\label{fig:class_conditional_arch}
	\end{figure*}
	
	\noindent \textbf{Element-wise powers.} Another straightforward choice for the function $f_{i}$ consists in applying element-wise powers so as to raise any input value $x$ to a power $i$ following:
	\begin{equation}
		f_{i}^{\mypow}(x) = x^{i}  
		\label{eq:pow_function_integer}
	\end{equation}
	Then by defining a vector of powers $\boldsymbol{\delta}  \in \mathbb{R}^{1 \times p}$ initialized with values from $1$ to $p$:
	\begin{equation}
		\mypowsymbol = \begin{bmatrix}
			1 & 2 & \cdots & p
		\end{bmatrix}
	\end{equation}
	the functional matrix expression $\Fc_{\y, p}^{\mypow}$ can be written as: 
	\begin{equation}
		\Fc_{\y, p}^{\mypow} = \frac{1}{p}\sum_{i=1}^{p}  \alpha_{i} f_{\mypowsymbol_i}^{\mypow}(\Ab_{\y} \Bb_{\y})
		\label{eq:fn_sum_pow_no_hyperparams}
	\end{equation}
	where the function $f_{\mypowsymbol_i}^{\mypow}$ is applied element-wise to the product $\Fb_{\y}$ = $\Ab_{\y} \Bb_{\y}$, and the rank of the final matrix $\Fc_{\y, p}^{\mypow}$ can also theoretically increase to $p$. By default the hyperparameters of $\mypowsymbol$ are frozen and not optimizable. However, for further expressiveness they may also be optimized using gradient descent, and when doing so, a surrogate formulation that works with decimal powers is employed \footnote{\textcolor{black}{We experimentally found that the formulation in Eq.~\ref{eq:pow_function_integer} is not defined for real numbers with negative values and decimal powers in PyTorch (it outputs NaN: Not a Number).}},  and defined as \begin{equation}
		f_{\mypowsymbol_i}^{\mypowsigned}(x) = \sign(x) \cdot (\abs(x))^{\mypowsymbol_i}
		\label{eq:pow_function_decimal_optimizable}
	\end{equation}
	where $\sign(.)$ and $\abs(.)$ respectively denote the sign and absolute value functions. 
	Under these conditions, the values in $\mypowsymbol$ will become decimal during training and the resulting functional matrix will be rewritten as:
	\begin{equation}
		\Fc_{\y, p}^{\mypowsignedlearned} = \frac{1}{p}\sum_{i=1}^{p}  \alphalora_{i} f_{\boldsymbol{\delta_i}}^{\mypowsigned}(\Ab_{\y} \Bb_{\y})
		\label{eq:fn_sum_pow_with_hyperparams}
	\end{equation}
	
	While using exponents can theoretically increase the rank up to $p$, our experiments will show that numerical approximations often yield a rank lower than $p$. This happens because higher powers may lead to infinitesimally small terms.  \\
	
	\noindent \textbf{Trigonometric functions.} Trigonometric functions, like the cosine, offer a powerful way to increase the rank of our functional matrices due to their orthogonality and periodicity. We can define an element-wise cosine function as:
	\begin{equation}
		f_{i}^{cos}(x) = \cos(\myfreqsymbol_i \cdot  x) 
	\end{equation}
	We also define a frequency vector $\myfreqsymbol \in \Rb^{1 \times p}$, initialized with values from $1$ to $p$.
	\begin{equation}
		\myfreqsymbol = \begin{bmatrix}
			\myfreqsymbol_1 & \myfreqsymbol_2 & \cdots & \myfreqsymbol_p
		\end{bmatrix}
	\end{equation}

	\noindent Hence, the function expression of the matrix $\Fc_{\y, p}^{\mycos}$ can be defined as:
	\begin{equation}
		\Fc_{\y, p}^{\mycos} = \frac{1}{p}\sum_{i=1}^{p}  \alphalora_{i} f_{\myfreqsymbol_i}^{\mycos}(\Ab_{\y} \Bb_{\y})
		\label{eq:fn_cos_no_hyperparams}
	\end{equation}
	Here $f_{\myfreqsymbol_i}^{\mycos}(\Ab_{\y} \Bb_{\y})$ is applied element-wise to the product of $\Ab_{\y} \Bb_{\y}$. By default, the $\myfreqsymbol$ hyperparameters are not optimized. However for better expressiveness, they can be optimized, in which case the functional matrix is assigned an $\myfreqsymbol$ superscript, and is rewritten as:
	\begin{equation}
		\Fc_{\y, p}^{\mycosfreq} = \frac{1}{p}\sum_{i=1}^{p}  \alphalora_{i} f_{\myfreqsymbol_i}^{\mycos}(\Ab_{\y} \Bb_{\y})
		\label{eq:fn_cos_with_hyperparams}
	\end{equation}
	\noindent A particularly interesting property of the cosine function, which our experiments will confirm, is its ability to increase a matrix's rank beyond $p$. This makes it potentially more expressive than other functional forms we have discussed. \\
		
	\noindent \textbf{Overall Visualization.} Figure \ref{fig:lora_feature_modulation_conv} illustrates the reparametrization of the convolutional weights in a single layer of the incremental generative model using the proposed functional LoRA. It can be noted that the functional matrix $\Fc_{\y}$ is reshaped to have dimensions in $\Rb^{C_{out} \times C_{in} \times 1 \times 1}$, so that the Hadamard product $\Wb_{0} \odot \Fc_{y} $ can be performed. Figure \ref{fig:class_conditional_arch} illustrates how convolutional layers are adapted at different layers of the generative model using the functional matrices $\Fc_{\y}$. 
	\begin{algorithm}[!h]
		\footnotesize 
		\KwIn{Dataset ${\Dc^{t}}$ for $t \in [1, T]$  \tcp*[r]{\scriptsize Image dataset}} 
		Initialize the incremental generative model $\incrmodel_{\incrmodelparams, \y}(.)$ \\
		\For{$t \gets 1$ to $T$} {			
			
			\If{$t == 1$}{
				Train $\incrmodel_{\incrmodelparams, \y}(.)$ on $\Dc^{1}$ without any constraints \tcp*[r]{\scriptsize Unconstrained training of all parameters}
				Initialize and train classifier $c_{\paramclassif}(.)$ on $\Dc^{t}$ \tcp*[r]{\scriptsize Classifier trained only on current task data}
			}
			\Else{
				\For{$\y \in \Yc^{t}$} {			
					Train all the matrices $\Ab_{\y}$ and $\Bb_{\y}$ (also the associated hyperparameters if applicable)  associated to $\y$ and used in the LoRA reparametrization for $\incrmodel_{\incrmodelparams, \y}(.)$ on $\Dc^{t}$  \tcp*[r]{\scriptsize LoRA training.}
				}
				Sample synthetic dataset $\Dc_{gen}^{1:t}$ with labels in $\Yc^{1:t}$ using $\incrmodel_{\incrmodelparams, \y}(.)$ \\
				Initialize and  train classifier $c_{\paramclassif}(.)$ on $\Dc_{gen}^{1:t}$ \tcp*[r]{\scriptsize Classifier trained only on synthetic data}
			}
			Evaluate accuracy of $c_{\paramclassif}(.)$ on the test set\\
		}
		\caption{Incremental training pipeline}
		\label{alg_incr_train_lora_diff}
	\end{algorithm}
	
	\noindent \textbf{Proposed Algorithm.} Algorithm \ref{alg_incr_train_lora_diff} shows the incremental training pipeline of our method. Most notably the generative model is trained without any constraint in the first task and without using any LoRA reparametrization. Similarly, the classifier is trained only on the real training data. Then in subsequent tasks, the generative model creates a synthetic dataset $\Dc_{gen}^{1:t}$ for all previous tasks with labels in $\Yc^{1:t}$ to ensure all classes have the same distribution. Finally, the classifier is trained on the fully synthetic dataset, and is then evaluated on the test set.

	\subsection{Layer Selection} In order to measure the importance of each neural network layer, we define a metric that quantifies the sensitivity of a given layer to a given task. Since all entries in the weight matrices $\Ab_{\y}$ and $\Bb_{\y}$ are initialized to 1, the importance of a given layer $l$ can be defined as
	\begin{equation}
		I_{l, \y} = \frac{1}{2} \left( \frac{ \lVert \Ab_{l, \y} - 1 \rVert_{1}}{\dim(\vecc(\Ab_{l, \y}))} + \frac{\lVert \Bb_{l, \y} - 1 \rVert_{1}}{\dim(\vecc(\Bb_{l, \y})|} \right)
		\label{eq:importance_layer}
	\end{equation}
	Here, $\lVert . \rVert_{1}$ denotes the L1 norm, $\vecc(.)$ represents the vectorization of a matrix, and $\dim(.)$ signifies the number of entries in a vector. $\Ab_{l, \y}$ and $\Bb_{l, \y}$  are the LoRA matrices for layer $l$ and class $\mathbf{y}$. This formulation evaluates the average L1 distance from the initial value of 1. To extend this importance to all classes, we average this measure as
	\begin{equation}
		I_{l} = \frac{1}{|\Yc|} \sum_{\y \in \Yc} I_{l, \y}
		\label{eq:importance_layer_class_avg}
	\end{equation}
	where $|\Yc|$ is the cardinal of the set of labels. \\

	\section{Experiments} \label{sec:params_dynamic}
	This section showcases the impact of the proposed incremental generative model, with efficient parameter allocation for each new class. We start by performing an ablation on the functions used to increase the rank, and then we compare it against closely related PEFT methods. Finally, we evaluate it with respect to closely related works involving incremental diffusion models trained from scratch \textcolor{black}{— and against} methods leveraging pre-trained stable diffusion models for continual learning.
	
	\subsection{Datasets and Evaluation Metrics}
	
	\noindent \textbf{Dataset.} We evaluate the proposed method on a wide range of challenging datasets commonly used in incremental learning with generative models, most notably:
	\begin{itemize}
		\item CIFAR10 \cite{cifar} which contains 50000 image for training and 10000 images for testing, equally split amongst 10 classes. This dataset is split into 5 equal tasks, each with an equal number of classes. This split will also be denoted as \emph{"2-2"} since the first task contains 2 classes, and the following ones also contain 2 classes.
		\item CIFAR100 \cite{cifar} which contains 50000 image for training and 10000 images for testing, equally split amongst 100 classes. This dataset is split into 5 or 10 equal tasks, each with an equal number of classes. This split will also be denoted as \emph{"20-20"} (resp \emph{"10-10"}) since the first task contains 20 classes (resp 10 classes), and the following ones also contain 20 classes (resp 10 classes).
		\item ImageNet100 \cite{imagenet} which contains 130000 images for training and 5000 images for testing equally split amongst 100 classes and resized to a dimension of $64\times64$. This dataset is split into 5 tasks, each with an equal number of classes. This split will also be denoted as \emph{"20-20"} since the first task contains 20 classes, and the following ones also contain 20 classes.
	\end{itemize}
	
	\noindent \textbf{Metrics.} We evaluate all methods using the average incremental accuracy (AIA) and the last accuracy (LA) of the classifier defined as following \cite{cil_survey_table, icarl}.  Given an accuracy score $A_{t}$ obtained on a test set $\Dc^{t}_{test}$, the average accuracy ($AA_{T}$)  up to an arbitrary task $T$ is formally defined as: 
		\begin{equation}
			AA_{T}= \frac{1}{T} \sum_{t=1}^{T} A_{t}
			\label{eq:aa_metric_continual}
		\end{equation}
		which corresponds to the mean of the accuracies achieved on all tasks encountered thus far. In particular, we also define the last accuracy (LA), as the last score $AA_{T}$ obtained after training on the last task. The average incremental accuracy (AIA) is defined as:
		\begin{equation}
			AIA_{T}= \frac{1}{T} \sum_{t=1}^{T} AA_{t}
			\label{eq:aia_metric_continual}
		\end{equation}
		which is the arithmetic mean of the average accuracies computed across all encountered tasks.  \\
	
	\noindent \textbf{Implementation Details:} During the first task, an Adam optimizer is used to train the flow matching model using a learning rate of $2.0 \times 10^{-4}$ (resp $1.0 \times 10^{-4}$), linearly warmed up for the first 1000 iterations, for a total of 500 epochs (resp 250 epochs) on CIFAR10-100 (resp ImageNet100). During this first task,  an exponential moving average (EMA) with a decay factor $\beta_{1} = 0.9995$ is also used starting from epoch 200 (resp 100) on CIFAR10-100 (resp ImageNet100). During subsequent tasks, the learning rate of the Adam optimizer is set to $1.0 \times 10^{-2}$, and the flow matching model is trained for 300  epochs (resp 200 epochs) on CIFAR10-100 (resp ImageNet100) using a decay factor $\beta_{1} = 0.995$ for the EMA, starting from epoch 100; and the batch size is 128 for all datasets. Concerning the classifier, we utilize a ResNet18 trained for 30 epochs and using a batch size of 100 on all datasets. The optimizer used is SGD with a learning rate of $10^{-3}$, a momentum of $0.9$, a weight decay of $5\cdot 10^{-4}$, and a one cycle scheduler \cite{one_cycle_lr} with a maximum learning rate of $0.1$. The augmentations used are random cropping and horizontal flip. Unless mentioned otherwise, the solver used when sampling the flow matching model is a dopri5 (Dormand-Prince) with error tolerances of $1.0 \times 10^{-4}$. In all experiments, we utilize either a single A100 or H100 GPU, with 80GB of RAM.

	\subsection{Baselines and Comparative Methods} \label{sec:comparative_methods_funlora}
	\noindent \textbf{Baseline.} For the baseline, we consider a LoRA applied for feature modulation on convolutional layers as expressed in Eq~\ref{eq:feature_modulation_vanilla_decomp} and \ref{eq:feature_modulation_vanilla_decomp_product} with a rank of 1. Optionally, the mathematical operation used may be a standard addition $\oplus$ (as in a vanilla LoRA) instead of the product $\otimes$, in which case the matrices $\Ab$ and $\Bb$ respectively follow a Kaiming initialization, and a 0 initialization.  \\
	\noindent \textbf{Comparative Method 1 - Scale and Shift.} For this comparative method, we consider learning class conditional coefficients for scaling and shifting feature maps in convolutional layers, as in \cite{scaling_and_shifting}. \textcolor{black}{This} can be considered a generalization of methods modifying biases \cite{bitfit}, or learning different normalization layers \cite{peft_normalization}. \\
	\textbf{Comparative Method 2 - LoRA on attention layers.} This comparative method is the standard LoRA decomposition applied to attention layers as in \cite{lora}. In practice, the rank used is larger than 1 for all layers in the generative model, and the matrices are also conditioned on class labels. \\
	\textbf{Comparative Method 3 - LoRA on Convolutional Layers v1.} This corresponds to a LoRA decomposition applied to convolutional layers, as implemented in the source code of the official LoRA  paper~\cite{lora}. Most notably, the bottleneck dimension $r_{conv} =  \kernelsize \times r$ also includes the kernel size as described in section \ref{sec:lora_reminder_with_conv}. \\
	\textbf{Comparative Method 4 - LoRA on Convolutional Layers v2}. This corresponds to a LoRA decomposition applied to convolutional layers as in \cite{kernel_modulation}, which corresponds to a bottleneck of $r_{conv} = 1$ as described in section \ref{sec:lora_reminder_with_conv}. This formulation for $\Ab \in \Rb^{C_{out} \times 1 }$ and $\Bb \in \Rb^{ 1 \times (C_{in} \cdot \kernelsize \cdot  \kernelsize)}$ requires a minimal number of extra parameters to learn different filter groups.

	\subsection{Results}
\begin{figure}[!htb]
		\includegraphics[width=1.0\linewidth]{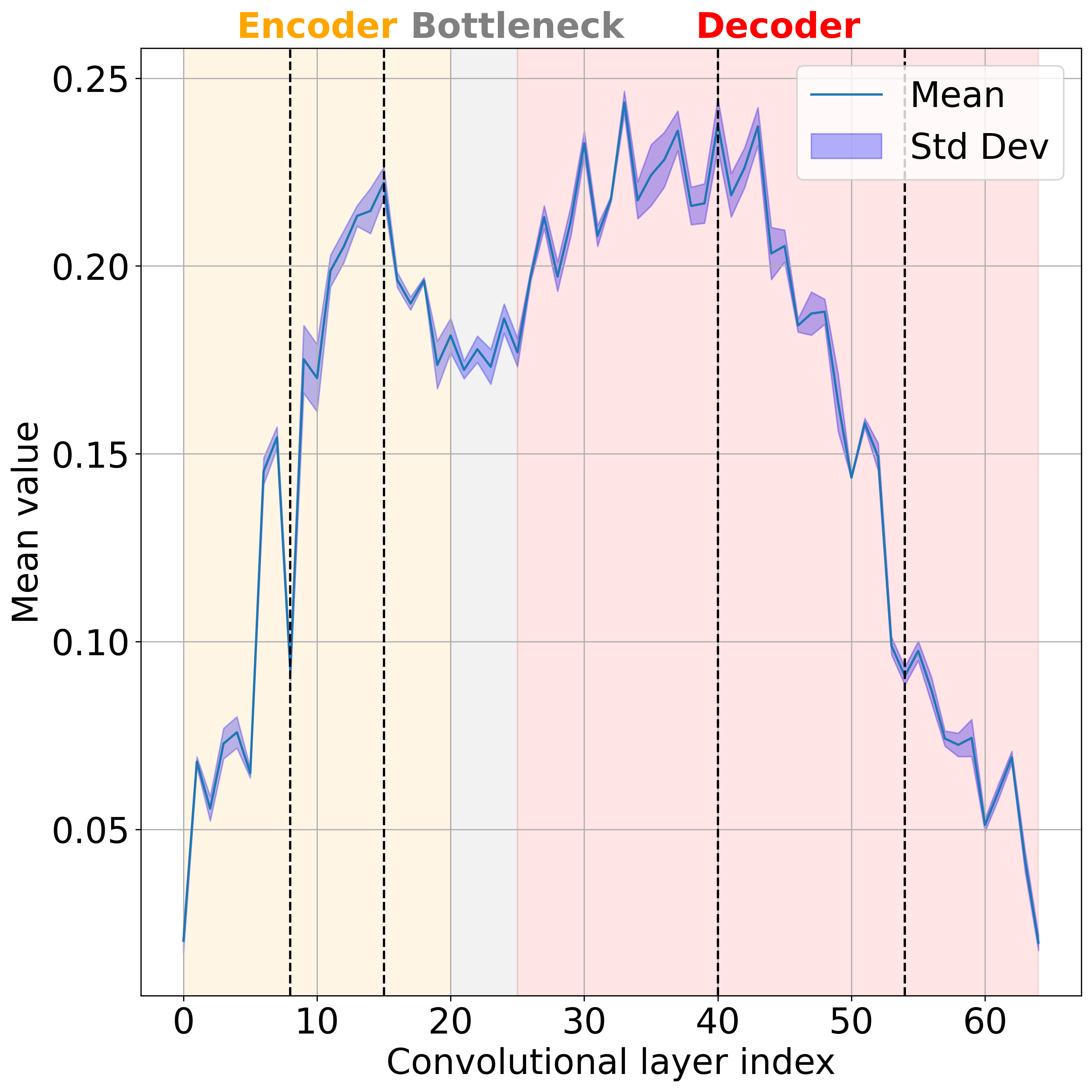}
		\centering
		\caption{Average importance $I_{l}$ (as in Eq~\ref{eq:importance_layer_class_avg}) calculated for each convolutional layer in the U-Net model, after incrementally training a flow matching model on CIFAR100. Layers with values that are furthest from 0 are the most important to adapt when learning a new class. Overall, the layers requiring less modifications are generally located next to the input and the output of the U-Net model.  Results are obtained on CIFAR100 with a split of 20-20.}
		\label{fig:plot_importances}
	\end{figure}

	\noindent \textbf{Layer Selection.} In order to develop a memory frugal PEFT approach, we first perform an analysis on the importance of each layer when learning new classes, so as to ignore redundant ones. Most notably, we utilize the \emph{"importance metrics"} defined in Eq~\ref{eq:importance_layer} ($I_{l, \y}$) and Eq~\ref{eq:importance_layer_class_avg} ($I_{l}$)  after training the generative model on the last task \textcolor{black}{— in order to} prioritize layers requiring the most significant parameter modification. To ensure those metrics are \emph{"unconstrained"} \textcolor{black}{by the choice of our functions (the cosine function can only have values between -1, and 1)},  we utilize the vanilla LoRA reparametrization  from Eq~\ref{eq:feature_modulation_vanilla_decomp} and Eq~\ref{eq:feature_modulation_vanilla_decomp_product}, and in Figure~\ref{fig:plot_importances} we show the values of $I_{l}$ for each layer in the neural network.  
	
	\begin{table}[!htb]
		\centering
		\resizebox{0.8\linewidth}{!}{%
			\begin{tabular}{cccc} 
				\toprule
				Indices                             & LA  $\uparrow$              & AIA  $\uparrow$            & PPC  $\downarrow$ \\ 
				\midrule
				0–64                                & $60.52 \pm 0.57$ & $68.64 \pm 0.84$ & 34.054K           \\
				5–54                                & $60.48 \pm 0.57$ & $68.39 \pm 0.76$ & 29.440K           \\
				15–44                               & $54.59 \pm 0.99$ & $64.16 \pm 0.73$ & 18.432K           \\
				15–54                               & $59.63 \pm 0.26$ & $67.84 \pm 0.96$ & 24.832K           \\
				25–44                               & $53.43 \pm 0.76$ & $63.71 \pm 0.63$ & 13.312K           \\
				25–54                               & $59.31 \pm 0.68$ & $67.70 \pm 0.98$ & 19.712K           \\
				35–54                               & $58.88 \pm 0.38$ & $67.39 \pm 0.86$ & 13.056K           \\
				\rowcolor[rgb]{1,0.867,0.867} 40–54 & $58.07 \pm 0.18$ & $66.99 \pm 1.02$ & 9.728K            \\
				45–54                               & $56.22 \pm 0.32$ & $65.70 \pm 1.25$ & 6.400K            \\
				40–49                               & $56.27 \pm 0.07$ & $65.57 \pm 0.94$ & 6.656K            \\
				45–49                               & $53.83 \pm 0.04$ & $64.01 \pm 0.99$ & 3.328K            \\
				\bottomrule
			\end{tabular}
		}
		\caption{Accuracy scores when adapting only a portion of the convolutional layers of our flow matching model based on their indices as shown in Fig~\ref{fig:plot_importances}. Results are obtained on CIFAR100 with a split of 20-20. Concerning the incremental tasks, the model is trained for 200 epochs (not 300 epochs which is the default for other experiments in this section)}
		\label{table:portion_adaptation}
	\end{table}
	
\begin{figure*}[!ht]
		\includegraphics[width=1.0\linewidth]{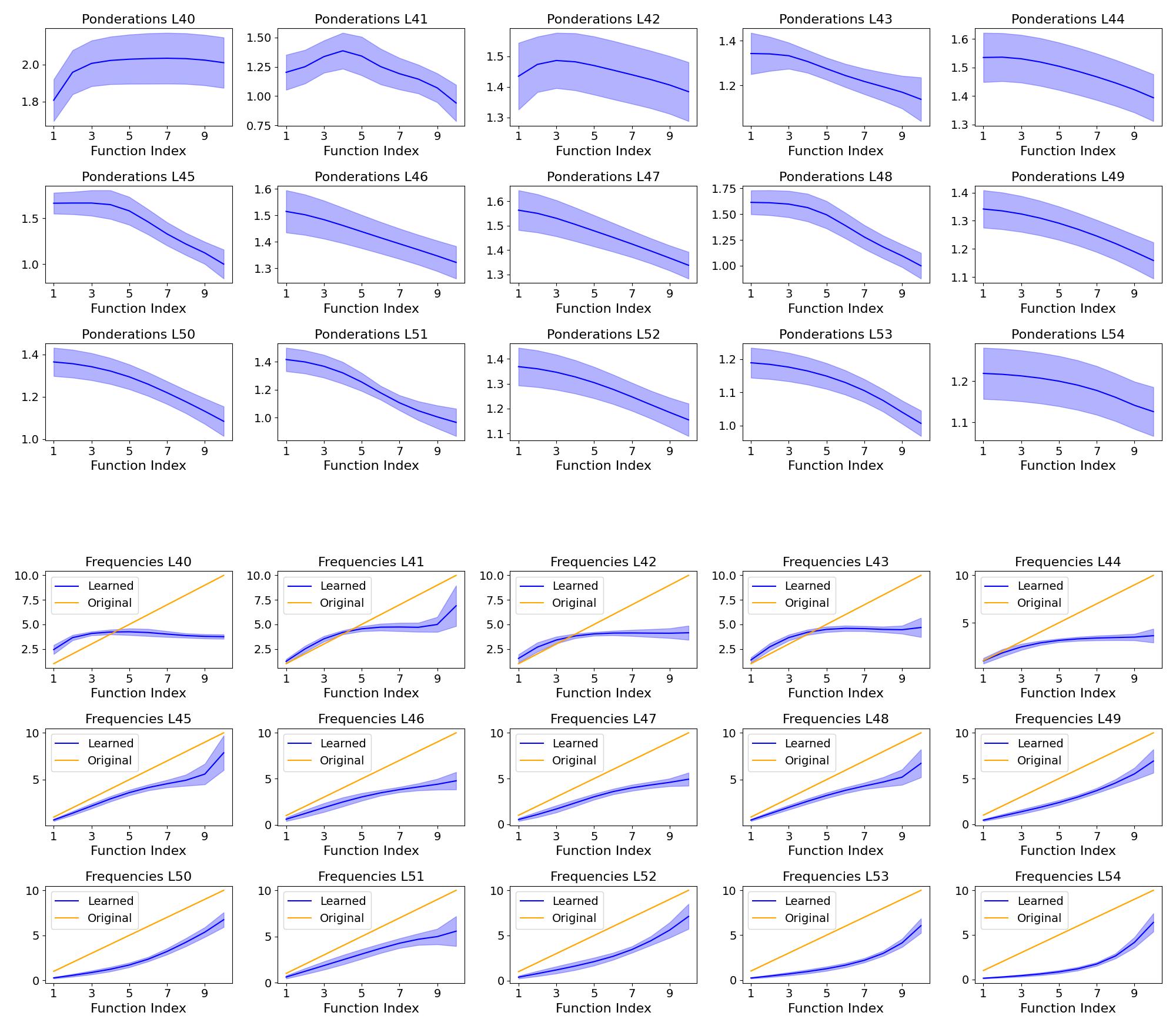}
		\centering
		\caption{Illustration of the ponderations values $\alphalora_{i}$, and the learned frequencies $\myfreqsymbol_i$ for each adapted layer after training our incremental flow matching model using the $\Fc_{\y, 10}^{\mycosfreq}$ reparametrization. Blue lines and shades represent the mean and standard deviation calculated on all classes.}
		\label{fig:plot_frequencies_lora_cos_10_funlora}
	\end{figure*}

		\begin{table*}[!htb]
		\centering
		\resizebox{0.8\linewidth}{!}{%
			\begin{tabular}{cclcccr} 
				\toprule
				\makecell{Function \\ Used} & \makecell{Train\\Hyperparams} & \multicolumn{1}{c}{\makecell{LoRA \\ Expression}}                 & \makecell{Layers \\ index}                & PPC $\downarrow$ & LA $\uparrow$                          & AIA  $\uparrow$                        \\ 
				\hline
				Vanilla add                 & NA                            & $\Wb_{0} \oplus \Fb_{\y}$                          & \multirow{7}{*}{\makecell{Index: 0--64}}  & 35.59K   & $\idf{59.23_{\pm 1.44}}$    & $\idf{67.17_{\pm 0.95}}$     \\
				Vanilla mul                 & NA                            & $\Wb_{0} \odot \Fb_{\y}$                           &                                           & 35.59K   & $\idf{60.84_{\pm 0.87}}$    & $\idf{68.82_{\pm 0.72}}$     \\
				Rshift                      & NA                            & $\Wb_{0} \odot \Fc_{\y, 10}^{\circshift}$          &                                           & 34.70K   & $\idf{58.55_{\pm 0.57}}$    & $\idf{67.19_{\pm 0.92}}$     \\
				Cosine                      & x                             & $\Wb_{0} \odot \Fc_{\y, 10}^{\mycos}$              &                                           & 34.70K   & $\idf{60.59_{\pm 0.53}}$    & $\idf{68.49_{\pm 0.73}}$     \\
				Cosine                      & v                             & $\Wb_{0} \odot \Fc_{\y, 10}^{\mycosfreq}$          &                                           & 35.35K   & $\mathbf{61.82_{\pm 0.56}}$ & $\mathbf{69.18_{\pm 0.93}}$  \\
				Power                       & x                             & $\Wb_{0} \odot \Fc_{\y, 10}^{\mypow}$              &                                           & 34.70K   & $\idf{61.33_{\pm 0.32}}$    & $\idf{68.90_{\pm 0.81}}$     \\
				Power                       & v                             & $\Wb_{0} \odot \Fc_{\y, 10}^{\mypowsignedlearned}$ &                                           & 35.35K   & $\idf{60.88_{\pm 0.66}}$    & $\idf{68.79_{\pm 0.72}}$     \\ 
				\hline\hline
				Vanilla add                 & NA                            & $\Wb_{0} \oplus \Fb_{\y}$                          & \multirow{7}{*}{\makecell{Index: 40--54}} & 11.26K   & $\idf{59.29_{\pm 0.34}}$    & $\idf{67.65_{\pm 1.11}}$     \\
				Vanilla mul                 & NA                            & $\Wb_{0} \odot \Fb_{\y}$                           &                                           & 11.26K   & $\idf{58.78_{\pm 0.52}}$    & $\idf{67.13_{\pm 1.07}}$     \\
				Rshift                      & NA                            & $\Wb_{0} \odot \Fc_{\y, 10}^{\circshift}$          &                                           & 9.88K    & $\idf{56.15_{\pm 0.66}}$    & $\idf{65.78_{\pm 0.95}}$     \\
				Cosine                      & x                             & $\Wb_{0} \odot \Fc_{\y, 10}^{\mycos}$              &                                           & 9.88K    & $\idf{60.01_{\pm 0.63}}$    & $\idf{68.01_{\pm 0.93}}$     \\
				Cosine                      & v                             & $\Wb_{0} \odot \Fc_{\y, 10}^{\mycosfreq}$          &                                           & 10.03K   & $\mathbf{60.07_{\pm 0.72}}$ & $\mathbf{68.06_{\pm 0.87}}$  \\
				Power                       & x                             & $\Wb_{0} \odot \Fc_{\y, 10}^{\mypow}$              &                                           & 9.88K    & $\idf{59.35_{\pm 0.35}}$    & $\idf{67.60_{\pm 0.95}}$      \\
				Power                       & v                             & $\Wb_{0} \odot \Fc_{\y, 10}^{\mypowsignedlearned}$ &                                           & 10.03K   & $\idf{58.70_{\pm 0.81}}$     & $\idf{67.41_{\pm 0.96}}$     \\
				\bottomrule
			\end{tabular}
		}
		\caption{This table presents a comparison and ablation study of various functions designed to increase the rank of matrices used in LoRA reparametrization. We analyze their performance when applied to all convolutional layers (upper part of the table) versus only the most important ones (lower part). The \emph{"vanilla add"} and \emph{"vanilla mul"} entries serve as baselines, representing standard LoRA applied to convolutional filters using rank-1 or rank-2 matrices; these baselines employ more parameters than our proposed methods. Specifically, "mul" signifies multiplication with the original matrix, while "add" refers to the standard addition common in LoRA. NA stands for not applicable, in cases when there are no hyperparameters for specific functions. Our proposed methods utilize 10 functions, and "train hyperparameters" indicates whether the frequencies $\myfreqsymbol$ (for the cosine function) or the power exponent (for the power function) are learnable. The specific LoRA expression for each method is detailed in a separate column. All results were obtained on the CIFAR100 dataset, with each task containing 20 classes (scenario 20-20). Across all tested cases, the cosine function consistently achieves the highest results and reliably benefits from trainable hyperparameters ($\Fc_{\y, 10}^{\mycosfreq}$), a trend not observed with the "power" function.}
		\label{table:ablation_fn_used}
	\end{table*}
	\begin{figure*}[!htb]
		\includegraphics[width=1.0\textwidth]{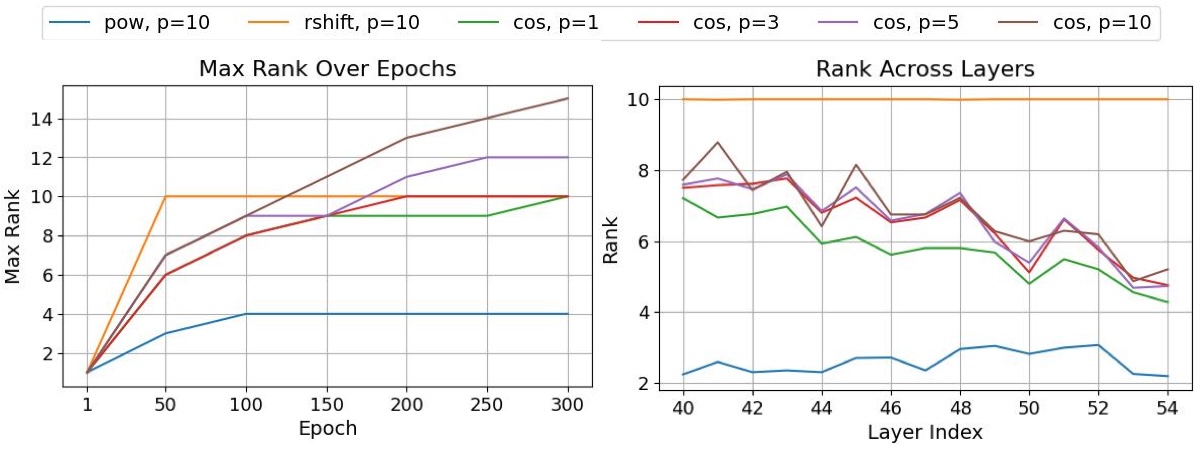}
		\centering
		\caption{ i) This table provides a comprehensive analysis of LoRA matrix characteristics: (i) the maximum LoRA matrix rank across all layers as the number of epochs increases in the left.  (ii) the average rank of the LoRA matrix across all layers (indices 40 to 54) is presented in the right. Notably, the "rshift-10" operation consistently increases the rank to 10, corresponding to the total number of distinct functions applied. Conversely, the "cosine" operation achieves the highest maximum rank, potentially allowing for greater localized expressiveness, whereas the "power" function only reaches a maximum rank of 4.}
		\label{fig:combined_plots_funlora}
	\end{figure*}
	It can be noted that the standard deviation between classes is very small, which indicates the U-Net architecture incorporates conditional information at very specific layers, regardless of the class label. Overall, the layers closest to the input and output of the U-Net model do not require any modifications, which suggests only noise is introduced in these portions. Furthermore, according to the importance metric $I_{l}$, the decoder part of the U-Net model is the most important to fine-tune, and this is also coherent since it is the closest to the final output. Besides the decoder part of the model, the encoder and the bottleneck also have high \emph{"importance"} values, suggesting middle layers of the U-Net are important to fine-tune. 
	\indent To further reduce the number of extra parameters when learning a new class, we progressively adapt fewer layers, specifically those with indices 40-54, as shown in Table~\ref{table:portion_adaptation}. \\
	Our analysis reveals that the decoder's middle portion (indices 45-54) is the most critical, with a significant 5-accuracy-point loss if not adapted, while the bottleneck (25-35) or encoder (0-25) only incur marginal decreases. This optimized approach offers the best trade-off between memory footprint and performance, using three times fewer parameters (9.728K) than fine-tuning all layers (0-64) while only losing 2 accuracy points. This finding appears somewhat counterintuitive when solely considering the $I_{l}$  metric in Figure~\ref{fig:plot_importances}, which suggests indices 25-44 are most important but surprisingly yield lower accuracy (of $53.43$) despite using more parameters. This can be explained by two main reasons: firstly, layers 40-54 include crucial skip connections from the encoder's most important part (indices 8-15), indirectly impacting encoder portions not targeted by adapting only 25-44; secondly, changes to layers 40-54 can absorb modifications that would otherwise occur in downstream layers (25-39), extending their impact across the entire encoder network and its skip connections. From the architectural point of view, indices 41-54 comprise layers 4 through 8 of the PyTorch (denoted as \emph{"output\_blocks.4"} to \emph{"output\_blocks.8"}) which scale with any diffusion U-Net size, while index 40 refers to a skip connection from previous block (denoted as "output blocks.3.").  Consequently, in the remainder of this section, only these most important convolutional layers (40-54) are adapted in the incremental U-Net model unless otherwise stated.\\
\noindent \textbf{Ablations.} In Table~\ref{table:ablation_fn_used}, we perform an ablation on various functions proposed in section~\ref{sec:lora_based_approach}, most notably the cosine function, with or without learning the hyperparameters (the frequencies) $\Fc_{\y, p}^{\mycos}$, $\Fc_{\y, p}^{\mycosfreq}$; the power functions with or without learning the hyperparameters (the decimal powers) $\Fc_{\y, p}^{\mypow}$, $\Fc_{\y, p}^{\mypowsignedlearned}$; and the right shift function $\Fc_{\y, p}^{\circshift}$ which does not have any hyperparameters. Those functions are also compared with vanilla LoRA reparametrization using matrices with rank 1 and 2.
	In all cases, it can be noted that the \emph{"cosine"} function achieves the highest results, and consistently benefits from trainable hyperparameters. Furthermore, it outperforms baselines with more parameters by 1 accuracy point. This expressiveness stems from the higher rank (higher than 1 while $\Ab$ and $\Bb$ are only of rank 1) obtained using the cosine reparametrization $\Fc_{\y, 10}^{\mycosfreq}$ as illustrated in Figure~\ref{fig:combined_plots_funlora}.  \\
	\noindent Additionally, when compared to other functions, the maximum rank achieved with the cosine can be higher than the number of functions $p$ (10 in this case), reaching a maximum value of 15. This is not the case for the power function reaching a lower maximum rank of 4, nor the right shift which always reaches a maximum rank of $p$. 
	\noindent In Figure~\ref{fig:plot_frequencies_lora_cos_10_funlora}, we further analyze the ponderations values $\alphalora_{i}$, and the learned frequencies $\myfreqsymbol_i$ for each adapted layer using the $\Fc_{\y, 10}^{\mycosfreq}$ reparametrization after training. One may note that the ponderation values generally decrease as the functional index increases. This means that frequencies initialized with higher values (which typically corresponds to noise), are given less importance in the overall sum. Regarding the learned frequencies in $\mycosfreq$ (which were initially the standard affine function shown by the orange line), we observe a different behavior across input, middle, and output layers. Most notably, input layers (indices 40-44) tend to learn similar frequencies starting from functional index 3, suggesting that just three functions capture sufficient expressiveness. As these functions correspond to mid-level frequencies, these layers (40-44) likely help incorporate class-conditional information into the generative model. The subsequent layers (45-54) generally resemble a dampened linear function, with the final layers (53-54) emphasizing only low frequencies. This suggests an impact on entire image extents, including background and foreground class information.

	\noindent Finally, Table~\ref{table:operation_used_lora} presents an ablation study on the various matrix operations usable with LoRA: addition (add), multiplication (mul), or multiplication and addition (mul-add) when employing the $\Fc_{\y, 10}^{\mycosfreq}$ reparametrization. Overall, the multiplication operation, which initializes matrices $\Ab_{\y}$ and $\Bb_{\y}$ with all entries as 1, yields the best results. Consequently, all subsequent results in this paper are presented using $\Fc_{\y, 10}^{\mycosfreq}$ with this specific initialization.
\begin{table*}[!htb]
	\centering
	\begin{minipage}{0.48\linewidth}
		\centering
		\resizebox{\linewidth}{!}{%
			\begin{tabular}{lcc}
				\toprule
				Operation & CIFAR10 & CIFAR100 \\
				\midrule
				Add: $\Wb_{0} \oplus \Fc_{\y, 10}^{\mycosfreq}$ & 18.76 & 14.56 \\
				Mul-Add: $\Wb_{0} \odot (1 \oplus \Fc_{\y, 10}^{\mycosfreq})$ & 83.98 & 59.12 \\
				Mul: $\Wb_{0} \odot \Fc_{\y, 10}^{\mycosfreq}$ & \textbf{84.02} & \textbf{60.07} \\
				\bottomrule
			\end{tabular}
		}
		\caption{This table compares the Last Accuracy (LA) score, achieved using various operations to adapt the weight $\Wb_{0}$ with $\Fc_{\y, p}^{\mycosfreq}$. For multiplication ("mul"), matrices $\Ab_{\y}$ and $\Bb_{\y}$ are initialized with all entries as 1, while other operations use the standard LoRA initialization (zero and Kaiming initialization). The dataset splits for CIFAR10 and CIFAR100 each involve 5 tasks with an equal number of classes (2-2 and 20-20 splits).}
		\label{table:operation_used_lora}
	\end{minipage}
	\hfill
	\begin{minipage}{0.48\linewidth}
		\centering
		\resizebox{\linewidth}{!}{%
			\begin{tabular}{ccccc}
				\toprule
				Ratio $k$ & LA $\uparrow$ & AIA $\uparrow$ & PPC $\downarrow$ & \makecell{Images\\per class\\equivalent} \\
				\midrule
				1  & $\idf{60.07 \pm 0.87}$ & $\idf{68.06 \pm 0.94}$ & 10.03K & 13 \\
				2  & $\idf{58.38 \pm 0.75}$ & $\idf{67.21 \pm 0.96}$ & 5.284K & 7  \\
				4  & $\idf{56.49 \pm 0.54}$ & $\idf{65.80 \pm 0.90}$ & 2.732K & 4  \\
				8  & $\idf{53.44 \pm 0.74}$ & $\idf{63.96 \pm 1.19}$ & 1.516K & 2  \\
				16 & $\idf{50.96 \pm 0.53}$ & $\idf{62.30 \pm 1.04}$ & 0.908K & 1  \\
				\bottomrule
			\end{tabular}
		}
		\caption{This table shows the Last Accuracy (LA) and Average Incremental Accuracy (AIA) when reusing LoRA parameters to save memory. "Images per class equivalent" indicates how many images could be stored in a memory buffer instead of allocating dynamic parameters. These results come from CIFAR100 split into 5 equal tasks.}
		\label{table:ratio_duplication_lora}
	\end{minipage}
\end{table*}

	\begin{table*}[!bp]
		\centering
		\resizebox{\linewidth}{!}{%
			\begin{tabular}{cccccccccc}
				\toprule
				\multirow{2}{*}{Method} & \multicolumn{3}{c}{\textbf{CIFAR10} 2-2} & \multicolumn{3}{c}{\textbf{CIFAR100} 20-20} & \multicolumn{3}{c}{\textbf{ImageNet100} 20-20} \\
				\cmidrule(lr){2-4} \cmidrule(lr){5-7} \cmidrule(lr){8-10}
				& LA$\uparrow$  & AIA$\uparrow$ & PPC$\downarrow$ & LA$\uparrow$ & AIA$\uparrow$ & PPC$\downarrow$ & LA$\uparrow$ & AIA$\uparrow$ & PPC$\downarrow$ \\
				\midrule
				Classifier Multitask & $\idf{93.88_{\pm 0.09}}$ & \multicolumn{2}{c}{NA} & $\idf{75.35_{\pm 0.02}}$ & \multicolumn{2}{c}{NA} & $\idf{76.55_{\pm 0.38}}$ & \multicolumn{2}{c}{NA} \\
				Generative Multitask & $\idf{90.56_{\pm 0.28}}$ & \multicolumn{2}{c}{NA} & $\idf{64.51_{\pm 0.13}}$ & \multicolumn{2}{c}{NA} & $\idf{62.95_{\pm 0.68}}$ & \multicolumn{2}{c}{NA} \\
				\hline \hline
				Vanilla conditioning only & $\idf{52.24_{\pm 4.09}}$ & $\idf{68.83_{\pm 3.00}}$ & 0.0K & $\idf{35.10_{\pm 0.51}}$ & $\idf{51.12_{\pm 0.83}}$ & 0.0K & $\idf{29.89_{\pm 0.77}}$ & $\idf{49.33_{\pm 0.73}}$ & 0.0K \\
				Scale and shift (Conv2d) \cite{scaling_and_shifting} & $\idf{82.85_{\pm 0.39}}$ & $\idf{88.30_{\pm 0.39}}$ & 11.78K & $\idf{58.18_{\pm 0.73}}$ & $\idf{66.91_{\pm 0.89}}$ & 11.78K & $\idf{56.64_{\pm 0.28}}$ & $\idf{67.75_{\pm 0.90}}$ & 16.13K \\
				LoRA attention layers \cite{lora} & $\idf{80.65_{\pm 1.40}}$ & $\idf{86.98_{\pm 0.52}}$ & 18.43K & $\idf{59.04_{\pm 0.50}}$ & $\idf{67.58_{\pm 0.97}}$ & 18.43K & $\idf{16.71_{\pm 0.45}}$ & $\idf{38.54_{\pm 0.68}}$ & 75.26K \\
				LoRA Conv2d v1 \cite{lora} & $\idf{82.80_{\pm 0.94}}$ & $\idf{88.31_{\pm 0.63}}$ & 12.29K & $\idf{57.45_{\pm 0.51}}$ & $\idf{66.31_{\pm 1.04}}$ & 12.29K & $\idf{55.87_{\pm 0.69}}$ & $\idf{67.15_{\pm 0.98}}$ & 14.98K \\
				LoRA Conv2d v2 \cite{kernel_modulation} & $\idf{82.15_{\pm 0.53}}$ & $\idf{87.73_{\pm 0.43}}$ & 10.75K & $\idf{58.71_{\pm 0.62}}$ & $\idf{67.33_{\pm 1.18}}$ & 10.75K & $\idf{57.46_{\pm 0.24}}$ & $\idf{68.15_{\pm 1.06}}$ & 14.21K \\
				FunLoRA:  $\Fc_{\y, 10}^{\mycosfreq}$ (ours) & $\mathbf{84.02_{\pm 0.44}}$ & $\mathbf{89.13_{\pm 0.33}}$ & \textbf{10.03K} & $\mathbf{60.07_{\pm 0.87}}$ & $\mathbf{68.06_{\pm 0.94}}$ & \textbf{10.03K} & $\mathbf{58.30_{\pm 0.60}}$ & $\mathbf{68.77_{\pm 0.61}}$ & \textbf{13.65K} \\
				\bottomrule
			\end{tabular}
		}
		\caption{Comparison against other PEFT methods including scaling and shifting feature maps, the vanilla LoRA applied on the attentions layers, and two LoRA methods applied on convolutional layers. We established our upper and lower bounds for comparison using various classifier accuracies. We considered: (i) the accuracy score of a classifier trained in a multitask setting (Classifier Multitask); (ii) the accuracy of a classifier trained in a multitask setting using synthetic data from a generative model also trained in a multitask setting (Generative Multitask); and (iii) the accuracy of a classifier trained in a multitask setting using synthetic data from a generative model trained with default conditioning (i.e., no LoRA reparametrization), which serves as the definitive lower bound.}
		\label{table:closest_comparative_lora_methods}
	\end{table*}

	\begin{table*}[!bp]
		\centering
		\resizebox{\linewidth}{!}{%
			\begin{tabular}{lcccccccc} 
				\toprule
				& CIFAR10                  & \multirow{2}{*}{P\#} & \multicolumn{2}{c}{CIFAR100}                        & \multirow{2}{*}{P\#} & ImageNet100              & \multirow{2}{*}{P\#} & \multirow{2}{*}{\begin{tabular}[c]{@{}c@{}}Pre-trained\\SD\end{tabular}}  \\ 
				\cmidrule(lr){2-2}\cmidrule(l){4-5}\cmidrule(lr){7-7}
				Dataset Split                                                        & 2-2                      &                      & 20-20                    & 10-10                    &                      & 20-20                    &                      &                                                                          \\ 
				\midrule
				DDGR \cite{ddgr}                                                     & $\idf{43.69_{\pm 2.60}}$ & 52.4M                & $\idf{28.11_{\pm 2.58}}$ & $\idf{15.99_{\pm 1.08}}$ & 52.4M                & $\idf{25.59_{\pm 2.29}}$ & 295M                 & \xmark                                                                   \\
				DGR Diffusion \cite{dgr}                                             & $\idf{59.00_{\pm 0.57}}$ & 52.4M                & $\idf{28.25_{\pm 0.22}}$ & $\idf{15.90_{\pm 1.01}}$ & 52.4M                & $\idf{23.92_{\pm 0.92}}$ & 295M                 & \xmark                                                                   \\
				GUIDE \cite{guide_diffusion}                               & $\idf{64.47_{\pm 0.45}}$ & 52.4M                & $\idf{41.66_{\pm 0.40}}$ & $\idf{26.13_{\pm 0.29}}$ & 52.4M                & $\idf{39.07_{\pm 1.37}}$ & 295M                 & \xmark                                                                   \\
				JDCL \cite{jdcl}                                                     & $\idf{83.69_{\pm 1.44}}$ & 52.4M                & $\idf{47.95_{\pm 0.61}}$ & $\idf{29.04_{\pm 0.41}}$ & 52.4M                & $\idf{54.53_{\pm 2.15}}$ & 295M                 & \xmark                                                                   \\
				DiffClass \cite{class_incremental_diffusion_pretraind_gaussian}      & NA                       & 983M                 & $\idf{62.21}$            & $\idf{58.40}$            & 983M                 & $\mathbf{67.26}$            & 983M                 & \cmark                                                                   \\ 
				\midrule
				FunLoRA: $\Fc_{\y, 10}^{\mycosfreq}$ (ours)                                                       & $\idf{84.02_{\pm 0.44}}$ & \textbf{35.84M}               & 
				$\idf{60.07_{\pm 0.87}}$ & $\idf{57.46_{\pm 0.55}}$                        & \textbf{36.61M}                & $\idf{58.30_{\pm 0.6}}$ & \textbf{90.69M}               & \xmark                                                                   \\
				\midrule
				\begin{tabular}[c]{@{}l@{}}FunLoRA:  $\Fc_{\y, 10}^{\mycosfreq}$ + \\ Resample (ours)\end{tabular} & $\mathbf{87.71_{\pm 0.36}}$                        & \textbf{35.84M}               & $\mathbf{67.89_{\pm 0.29}}$                        & $\mathbf{64.55_{\pm 0.6}}$                        & \textbf{36.61M}               & $\idf{63.03 _{\pm 0.69}}$ & \textbf{90.69M}              & \xmark                                                                   \\
				\bottomrule
			\end{tabular}
		}
		\caption{Comparison between the LA of our proposed reparametrization $\Fc_{\y, 10}^{\mycosfreq}$  applied to flow matching, against other closest related works using diffusion models for continual learning. P\# stands for the total number of parameters in the generative model. The works in \cite{ddgr} \cite{guide_diffusion} \cite{jdcl} \cite{dgr} train a similar U-Net backbone from scratch using more parameters, whereas \cite{class_incremental_diffusion_pretraind_gaussian} leverages a pre-trained stable diffusion (SD) model. \textcolor{black}{It should be noted that the scores in \cite{ddgr} and \cite{dgr} are taken from \cite{guide_diffusion}, which uses the same incremental splits as our experiments.} For the FunLoRA + Resample method, the dataset size is  increased by 10 and 5 on  respectively CIFAR10-100 and ImageNet100, and the solver used is RK4 with 5 NFE. This amounts to sampling time of 50 minutes on CIFAR10-100 (for a total of 500000 images), and 350 minutes on ImageNet100 (for a total of 650000 images) using a single A100 GPU with 80GB RAM.}
		\label{table:final_results_table_lora_avg_incr_acc}
	\end{table*}
	
	\begin{table*}[!htb]
		\centering
		\resizebox{0.8\linewidth}{!}{%
			\begin{tabular}{cccccccc}
				\toprule
				& \multicolumn{7}{c}{Solver = RK4} \\
				\cmidrule(lr){2-8}
				\makecell{Dataset size\\scaling factor} 
				& NFE=2 & NFE=3 & NFE=4 & NFE=5 & NFE=6 & NFE=10 & NFE=20 \\
				\midrule
				1  & 56.27 & 57.96 & 58.79 & 59.23 & 58.72 & 59.99 & 59.42 \\
				2  & 60.01 & 62.81 & 63.11 & 63.73 & 63.37 & 63.34 & 63.15 \\
				3  & 61.61 & 63.65 & 64.91 & 64.72 & 64.82 & 64.96 & 65.14 \\
				4  & 62.49 & 65.04 & 65.59 & 65.87 & 65.58 & 65.45 & 65.43 \\
				5  & 63.00 & 65.85 & 66.25 & 66.26 & 66.73 & 66.77 & 66.34 \\
				6  & 63.56 & 65.92 & 66.09 & 66.47 & 66.33 & 66.40 & 66.60 \\
				10 & 64.79 & 67.25 & 67.90 & 67.57 & 68.00 & 67.83 & 67.63 \\
				\midrule
				\makecell{Sampling time \\ (scaling factor = 1)} 
				& 115   & 229   & 344   & 459   & 576   & 689   & 1148 \\
				\bottomrule
			\end{tabular}
		}
		\caption{This table illustrates the influence on image quality and sampling speed when using a fixed number of function evaluations (NFE) with a fourth-order Runge-Kutta solver (RK4). It can be observed that unlike the DOPRI5 (Dormand-Prince) solver used previously, which required 1600 seconds for sampling, and reached 59.06 accuracy for the same seed — it is possible to obtain similar results using only 5 function evaluations (with a RK4 solver) which only required 459 seconds $\approx$ 8 minutes. The dataset size scaling factor indicates how many times larger the sampled dataset is compared to the original. Sampling times, measured in seconds on an A100 GPU with 80GB of RAM, are reported using a dataset scaling factor of 1. These results were obtained on CIFAR100 with 5 equally split tasks.}
		\label{table:nfe}
	\end{table*}

	\noindent  \textbf{Reduced Parameters.} Table~\ref{table:ratio_duplication_lora} examines the performance of the $\Fc_{\y, 10}^{\mycosfreq}$ reparametrization when using $k$ times fewer parameters for matrices $\Ab_{\y}$ and $\Bb_{\y}$. This is obtained by dividing the dimension of $\Ab_{\y}$ and $\Bb_{\y}$ by a ratio of $k$, which then requires duplicating values in the $\Fc_{\y, 10}^{\mycosfreq}$ matrix to match the dimensions of the initial weight matrix $\Wb_{0}$. Notably, even with parameters reduced by an order of magnitude, the AIA score decreases by a mere 6 accuracy points, clearly demonstrating the proposed method's \textbf{robustness}. Furthermore, highly competitive results are obtained by adding a number of parameters equivalent to storing just a single image in a memory buffer. \\
	\noindent  \textbf{Extra Comparisons.} In Table~\ref{table:closest_comparative_lora_methods}, we benchmark our proposed $\Fc_{\y, 10}^{\mycosfreq}$ reparametrization against other comparative methods detailed in Section~\ref{sec:comparative_methods_funlora}, including scaling and shifting feature maps, vanilla LoRA applied to attention layers, and two LoRA methods implemented on convolutional layers. To provide comprehensive context, we also include upper and lower bounds: the upper bound being the accuracy of a classifier trained in a multitask setting (Classifier Multitask) \textcolor{black}{— and the lower} bounds comprising the accuracy of a classifier trained in a multitask setting using synthetic data from a generative model also trained in a multitask setting (Generative Multitask). \textcolor{black}{We also provide} the accuracy of a classifier trained in a multitask setting using synthetic data from a generative model trained with default conditioning \textcolor{black}{(i.e., no extra LoRA reparametrizations are used on either the convolutional layers nor the attention layers for enhanced conditioning)}. Across all benchmarks, our proposed cosine reparametrization consistently achieves the highest results, improving performance by up to 1 extra accuracy point while using an equivalent number of parameters per class. Furthermore, the Last Accuracy (LA), closely approaches the upper bound on both CIFAR100 and ImageNet100. This clearly demonstrates the high efficiency of our proposed PEFT method. \\
	\noindent  \textbf{Final Comparison.} In Table~\ref{table:final_results_table_lora_avg_incr_acc}, we compare the AIA of our proposed reparametrization $\Fc_{\y, 10}^{\mycosfreq}$  against closest related works using diffusion models for continual learning. The works in \cite{ddgr} \cite{guide_diffusion}  \cite{jdcl} \cite{dgr} train a similar U-Net backbone (with more parameters) from scratch, whereas \cite{class_incremental_diffusion_pretraind_gaussian} leverages a pre-trained stable diffusion model (more details are available in the caption of Table~\ref{table:final_results_table_lora_avg_incr_acc}). It can be noted that our method consistently outperforms comparative methods on all benchmarks, and even by 20 accuracy points on CIFAR100 --- while using much less parameters for the backbone: respectively $36.61M$ vs $52.4M$ on CIFAR100, and $90.69M$ vs $295M$ on ImageNet100. Furthermore, the extra parameters introduced when learning those new classes with our method is very small: $0.802M$ parameters on CIFAR100 and $1.09M$ parameters on ImageNet100, which is negligible when compared to the number of parameters in the initial backbones of respectively $35.80M$ and $89.59M$. In fact, even after learning 1000 extra classes, our generative models would still have less parameters than these competing methods. Regarding comparisons with methods relying on pre-trained stable diffusion, we significantly outperform them on CIFAR100 when sampling more images than in the initial dataset (67.89 vs. 62.21). This is a major achievement, especially considering our model is trained on orders of magnitude fewer images. Our model is also significantly faster at sampling and adapting to new tasks compared to large pre-trained stable diffusion models, a crucial advantage for continual learning. This efficiency extends to vanilla diffusion models as well; they typically require more function evaluations (e.g., 100 on CIFAR100 as in \cite{guide_diffusion}) than flow matching, which can achieve decent results with only 2 NFE, as shown in Table~\ref{table:nfe}. We also share better results (available in Section~\ref{results_with_pretraining_on_coco} of the supplementary material), by pre-training the generative backbone on COCO dataset \cite{cocodataset}, in an unsupervised setting. This suggests that pre-training on data with similar distribution greatly enhances the results.
	\section{Conclusion}
	\noindent In this paper, we proposed a novel LoRA based PEFT method — dubbed FunLoRA — to adapt conditional generative model for continual classification. Most notably, our method leverages a frugal low rank reparametrized matrix (of rank 1), whose rank is functionally increased by applying carefully selected element-wise functions. As extensively corroborated through numerous experiments, the proposed method significantly outperforms related works in terms of accuracy, while requiring much less parameters.  Indeed, our proposed selection of important layers ensures only a portion of the backbone is adapted while maintaining competitive performance, and ensuring there is no forgetting.  Furthermore it can outperform methods using large pre-trained models such as stable diffusion, while requiring significantly less resources (compute power and data). This proposed method was also further bolstered by the usage of the novel flow matching paradigm, resulting in an accelerated sampling time with respect to diffusion models — and which holds tremendous potential in continual learning.

	\section*{Acknowledgements}
	This work was granted access to the HPC resources of IDRIS under the allocation 2025-AD011013954R2 made by GENCI.

	\pagebreak
	\clearpage
	
	\bibliographystyle{IEEEtran}
	\bibliography{refs/sota_1, refs/incremental, refs/activations, refs/augm, refs/datasets, refs/dl, refs/gaussian_cond, refs/gen_models, refs/cl}

\begin{thebibliography}{100}
\providecommand{\url}[1]{#1}
\csname url@samestyle\endcsname
\providecommand{\newblock}{\relax}
\providecommand{\bibinfo}[2]{#2}
\providecommand{\BIBentrySTDinterwordspacing}{\spaceskip=0pt\relax}
\providecommand{\BIBentryALTinterwordstretchfactor}{4}
\providecommand{\BIBentryALTinterwordspacing}{\spaceskip=\fontdimen2\font plus
\BIBentryALTinterwordstretchfactor\fontdimen3\font minus
  \fontdimen4\font\relax}
\providecommand{\BIBforeignlanguage}[2]{{%
\expandafter\ifx\csname l@#1\endcsname\relax
\typeout{** WARNING: IEEEtran.bst: No hyphenation pattern has been}%
\typeout{** loaded for the language `#1'. Using the pattern for}%
\typeout{** the default language instead.}%
\else
\language=\csname l@#1\endcsname
\fi
#2}}
\providecommand{\BIBdecl}{\relax}
\BIBdecl

\bibitem{mccloskey_catastrophic_forgetting}
\BIBentryALTinterwordspacing
M.~McCloskey and N.~J. Cohen, ``Catastrophic interference in connectionist
  networks: The sequential learning problem,'' ser. Psychology of Learning and
  Motivation, G.~H. Bower, Ed.\hskip 1em plus 0.5em minus 0.4em\relax Academic
  Press, 1989, vol.~24, pp. 109--165. [Online]. Available:
  \url{https://www.sciencedirect.com/science/article/pii/S0079742108605368}
\BIBentrySTDinterwordspacing

\bibitem{thrun1995lifelong}
S.~Thrun and T.~M. Mitchell, ``Lifelong robot learning,'' \emph{Robotics and
  autonomous systems}, vol.~15, no. 1-2, pp. 25--46, 1995.

\bibitem{ring1997child}
M.~B. Ring, ``Child: A first step towards continual learning,'' \emph{Machine
  Learning}, vol.~28, no.~1, pp. 77--104, 1997.

\bibitem{parisi2019continual}
G.~I. Parisi, R.~Kemker, J.~L. Part, C.~Kanan, and S.~Wermter, ``Continual
  lifelong learning with neural networks: A review,'' \emph{Neural networks},
  vol. 113, pp. 54--71, 2019.

\bibitem{chatgpt3}
T.~Brown, B.~Mann, N.~Ryder, M.~Subbiah, J.~D. Kaplan, P.~Dhariwal,
  A.~Neelakantan, P.~Shyam, G.~Sastry, A.~Askell \emph{et~al.}, ``Language
  models are few-shot learners,'' \emph{Advances in neural information
  processing systems}, vol.~33, pp. 1877--1901, 2020.

\bibitem{stable_diffusion}
R.~Rombach, A.~Blattmann, D.~Lorenz, P.~Esser, and B.~Ommer, ``High-resolution
  image synthesis with latent diffusion models,'' in \emph{Proceedings of the
  IEEE/CVF conference on computer vision and pattern recognition}, 2022, pp.
  10\,684--10\,695.

\bibitem{llama}
H.~Touvron, T.~Lavril, G.~Izacard, X.~Martinet, M.-A. Lachaux, T.~Lacroix,
  B.~Rozi{\`e}re, N.~Goyal, E.~Hambro, F.~Azhar \emph{et~al.}, ``Llama: Open
  and efficient foundation language models,'' \emph{arXiv preprint
  arXiv:2302.13971}, 2023.

\bibitem{lora}
E.~J. Hu, Y.~Shen, P.~Wallis, Z.~Allen-Zhu, Y.~Li, S.~Wang, L.~Wang, and
  W.~Chen, ``Lora: Low-rank adaptation of large language models,'' \emph{arXiv
  preprint arXiv:2106.09685}, 2021.

\bibitem{adapters}
N.~Houlsby, A.~Giurgiu, S.~Jastrzebski, B.~Morrone, Q.~De~Laroussilhe,
  A.~Gesmundo, M.~Attariyan, and S.~Gelly, ``Parameter-efficient transfer
  learning for nlp,'' in \emph{International conference on machine
  learning}.\hskip 1em plus 0.5em minus 0.4em\relax PMLR, 2019, pp. 2790--2799.

\bibitem{prompt_tuning}
B.~Lester, R.~Al-Rfou, and N.~Constant, ``The power of scale for
  parameter-efficient prompt tuning,'' \emph{arXiv preprint arXiv:2104.08691},
  2021.

\bibitem{peft_survey}
Z.~Han, C.~Gao, J.~Liu, J.~Zhang, and S.~Q. Zhang, ``Parameter-efficient
  fine-tuning for large models: A comprehensive survey,'' \emph{arXiv preprint
  arXiv:2403.14608}, 2024.

\bibitem{attention}
A.~Vaswani, ``Attention is all you need,'' \emph{Advances in Neural Information
  Processing Systems}, 2017.

\bibitem{peft_conv}
W.~Chen, Z.~Miao, and Q.~Qiu, ``Large convolutional model tuning via filter
  subspace,'' \emph{arXiv preprint arXiv:2403.00269}, 2024.

\bibitem{vpt}
M.~Jia, L.~Tang, B.-C. Chen, C.~Cardie, S.~Belongie, B.~Hariharan, and S.-N.
  Lim, ``Visual prompt tuning,'' in \emph{European conference on computer
  vision}.\hskip 1em plus 0.5em minus 0.4em\relax Springer, 2022, pp. 709--727.

\bibitem{lora_clr}
E.~Simsar, T.~Hofmann, F.~Tombari, and P.~Yanardag, ``Loraclr: Contrastive
  adaptation for customization of diffusion models,'' \emph{arXiv preprint
  arXiv:2412.09622}, 2024.

\bibitem{icarl}
S.-A. Rebuffi, A.~Kolesnikov, G.~Sperl, and C.~H. Lampert, ``icarl: Incremental
  classifier and representation learning,'' in \emph{Proceedings of the IEEE
  conference on Computer Vision and Pattern Recognition}, 2017, pp. 2001--2010.

\bibitem{dytox}
A.~Douillard, A.~Ram{\'e}, G.~Couairon, and M.~Cord, ``Dytox: Transformers for
  continual learning with dynamic token expansion,'' in \emph{Proceedings of
  the IEEE/CVF conference on computer vision and pattern recognition}, 2022,
  pp. 9285--9295.

\bibitem{lecun_cnn}
Y.~LeCun, B.~Boser, J.~S. Denker, D.~Henderson, R.~E. Howard, W.~Hubbard, and
  L.~D. Jackel, ``Backpropagation applied to handwritten zip code
  recognition,'' \emph{Neural Computation}, vol.~1, no.~4, pp. 541--551, 1989.

\bibitem{vit}
D.~Alexey, ``An image is worth 16x16 words: Transformers for image recognition
  at scale,'' \emph{arXiv preprint arXiv: 2010.11929}, 2020.

\bibitem{online_continual_learing_embedded}
T.~L. Hayes and C.~Kanan, ``Online continual learning for embedded devices,''
  \emph{arXiv preprint arXiv:2203.10681}, 2022.

\bibitem{tiny_ml_continual_learning}
L.~Ravaglia, M.~Rusci, D.~Nadalini, A.~Capotondi, F.~Conti, and L.~Benini, ``A
  tinyml platform for on-device continual learning with quantized latent
  replays,'' \emph{IEEE Journal on Emerging and Selected Topics in Circuits and
  Systems}, vol.~11, no.~4, pp. 789--802, 2021.

\bibitem{privacy_cl_1}
T.~Verma, L.~Jin, J.~Zhou, J.~Huang, M.~Tan, B.~C.~M. Choong, T.~F. Tan,
  F.~Gao, X.~Xu, D.~S. Ting, and Y.~Liu,
  ``\BIBforeignlanguage{en}{Privacy-preserving continual learning methods for
  medical image classification: a comparative analysis},''
  \emph{\BIBforeignlanguage{en}{Front. Med. (Lausanne)}}, vol.~10, p. 1227515,
  Aug. 2023.

\bibitem{privacy_cl_2}
E.~Verwimp, R.~Aljundi, S.~Ben-David, M.~Bethge, A.~Cossu, A.~Gepperth, T.~L.
  Hayes, E.~H{\"u}llermeier, C.~Kanan, D.~Kudithipudi \emph{et~al.},
  ``Continual learning: Applications and the road forward,'' \emph{arXiv
  preprint arXiv:2311.11908}, 2023.

\bibitem{fetril}
G.~Petit, A.~Popescu, H.~Schindler, D.~Picard, and B.~Delezoide, ``Fetril:
  Feature translation for exemplar-free class-incremental learning,'' in
  \emph{Proceedings of the IEEE/CVF winter conference on applications of
  computer vision}, 2023, pp. 3911--3920.

\bibitem{plastil}
\BIBentryALTinterwordspacing
G.~Petit, A.~Popescu, E.~Belouadah, D.~Picard, and B.~Delezoide, ``Plastil:
  Plastic and stable exemplar-free class-incremental learning,'' in
  \emph{Proceedings of The 2nd Conference on Lifelong Learning Agents}, ser.
  Proceedings of Machine Learning Research, S.~Chandar, R.~Pascanu, H.~Sedghi,
  and D.~Precup, Eds., vol. 232.\hskip 1em plus 0.5em minus 0.4em\relax PMLR,
  22--25 Aug 2023, pp. 399--414. [Online]. Available:
  \url{https://proceedings.mlr.press/v232/petit23a.html}
\BIBentrySTDinterwordspacing

\bibitem{learning_without_forgetting}
Z.~Li and D.~Hoiem, ``Learning without forgetting,'' \emph{IEEE transactions on
  pattern analysis and machine intelligence}, vol.~40, no.~12, pp. 2935--2947,
  2017.

\bibitem{variational_continual_learning}
C.~V. Nguyen, Y.~Li, T.~D. Bui, and R.~E. Turner, ``Variational continual
  learning,'' in \emph{International Conference on Learning Representations},
  2018.

\bibitem{continual_learning_in_gans}
A.~Seff, A.~Beatson, D.~Suo, and H.~Liu, ``Continual learning in generative
  adversarial nets,'' \emph{arXiv preprint arXiv:1705.08395}, 2017.

\bibitem{continual_learning_diffusion_distillation}
S.~Masip, P.~Rodriguez, T.~Tuytelaars, and G.~M. van~de Ven, ``Continual
  learning of diffusion models with generative distillation,'' \emph{arXiv
  preprint arXiv:2311.14028}, 2023.

\bibitem{mergan}
H.~Shin, J.~K. Lee, J.~Kim, and J.~Kim, ``Continual learning with deep
  generative replay,'' \emph{Advances in neural information processing
  systems}, vol.~30, 2017.

\bibitem{pseudo_rehearsal}
J.~Pomponi, S.~Scardapane, and A.~Uncini, ``Pseudo-rehearsal for continual
  learning with normalizing flows,'' \emph{arXiv preprint arXiv:2007.02443},
  2020.

\bibitem{ddgr}
\BIBentryALTinterwordspacing
R.~Gao and W.~Liu, ``{DDGR}: Continual learning with deep diffusion-based
  generative replay,'' in \emph{Proceedings of the 40th International
  Conference on Machine Learning}, ser. Proceedings of Machine Learning
  Research, A.~Krause, E.~Brunskill, K.~Cho, B.~Engelhardt, S.~Sabato, and
  J.~Scarlett, Eds., vol. 202.\hskip 1em plus 0.5em minus 0.4em\relax PMLR,
  23--29 Jul 2023, pp. 10\,744--10\,763. [Online]. Available:
  \url{https://proceedings.mlr.press/v202/gao23e.html}
\BIBentrySTDinterwordspacing

\bibitem{gppdm}
M.~M. Rahman, Q.~Mamun, and J.~Wu, ``Privacy-preserving spatial crowdsourcing
  in smart cities using federated and incremental learning approach,'' in
  \emph{2024 IEEE 100th Vehicular Technology Conference (VTC2024-Fall)}.\hskip
  1em plus 0.5em minus 0.4em\relax IEEE, 2024, pp. 1--7.

\bibitem{guide_diffusion}
B.~Cywi{\'n}ski, K.~Deja, T.~Trzci{\'n}ski, B.~Twardowski, and
  {\L}.~Kuci{\'n}ski, ``Guide: Guidance-based incremental learning with
  diffusion models,'' \emph{arXiv preprint arXiv:2403.03938}, 2024.

\bibitem{jdcl}
P.~Skier{\'s} and K.~Deja, ``Joint diffusion models in continual learning,''
  \emph{arXiv preprint arXiv:2411.08224}, 2024.

\bibitem{forgetting_flowers}
G.~Martinez, L.~Watson, P.~Reviriego, J.~A. Hernandez, M.~Juarez, and
  R.~Sarkar, ``Towards understanding the interplay of generative artificial
  intelligence and the internet,'' in \emph{International Workshop on Epistemic
  Uncertainty in Artificial Intelligence}.\hskip 1em plus 0.5em minus
  0.4em\relax Springer, 2023, pp. 59--73.

\bibitem{on_the_stability_of_iterative_retraining}
\BIBentryALTinterwordspacing
Q.~Bertrand, A.~J. Bose, A.~Duplessis, M.~Jiralerspong, and G.~Gidel, ``On the
  stability of iterative retraining of generative models on their own data,''
  in \emph{The Twelfth International Conference on Learning Representations,
  {ICLR} 2024, Vienna, Austria, May 7-11, 2024}.\hskip 1em plus 0.5em minus
  0.4em\relax OpenReview.net, 2024. [Online]. Available:
  \url{https://openreview.net/forum?id=JORAfH2xFd}
\BIBentrySTDinterwordspacing

\bibitem{self-consuming_generative_models_go_mad}
\BIBentryALTinterwordspacing
S.~Alemohammad, J.~Casco{-}Rodriguez, L.~Luzi, A.~I. Humayun, H.~Babaei,
  D.~LeJeune, A.~Siahkoohi, and R.~G. Baraniuk, ``Self-consuming generative
  models go {MAD},'' in \emph{The Twelfth International Conference on Learning
  Representations, {ICLR} 2024, Vienna, Austria, May 7-11, 2024}.\hskip 1em
  plus 0.5em minus 0.4em\relax OpenReview.net, 2024. [Online]. Available:
  \url{https://openreview.net/forum?id=ShjMHfmPs0}
\BIBentrySTDinterwordspacing

\bibitem{class_incremental_diffusion_pretraind_gaussian}
Z.~Meng, J.~Zhang, C.~Yang, Z.~Zhan, P.~Zhao, and Y.~Wang, ``Diffclass:
  Diffusion-based class incremental learning,'' in \emph{European Conference on
  Computer Vision}.\hskip 1em plus 0.5em minus 0.4em\relax Springer, 2024, pp.
  142--159.

\bibitem{diffusion_og}
J.~Sohl-Dickstein, E.~Weiss, N.~Maheswaranathan, and S.~Ganguli, ``Deep
  unsupervised learning using nonequilibrium thermodynamics,'' in
  \emph{International conference on machine learning}.\hskip 1em plus 0.5em
  minus 0.4em\relax PMLR, 2015, pp. 2256--2265.

\bibitem{class_incremental_learning_diffusion_pretrained}
Q.~Jodelet, X.~Liu, Y.~J. Phua, and T.~Murata, ``Class-incremental learning
  using diffusion model for distillation and replay,'' in \emph{Proceedings of
  the IEEE/CVF International Conference on Computer Vision}, 2023, pp.
  3425--3433.

\bibitem{reality_check_pretraining_popescu}
E.~Feillet, A.~Popescu, and C.~Hudelot, ``A reality check on pre-training for
  exemplar-free class-incremental learning,'' in \emph{2025 IEEE/CVF Winter
  Conference on Applications of Computer Vision (WACV)}, 2025, pp. 7625--7636.

\bibitem{c_lora_diffusion}
J.~S. Smith, Y.-C. Hsu, L.~Zhang, T.~Hua, Z.~Kira, Y.~Shen, and H.~Jin,
  ``Continual diffusion: Continual customization of text-to-image diffusion
  with c-lora,'' \emph{arXiv preprint arXiv:2304.06027}, 2023.

\bibitem{lora_continual_personalization_diffusion}
{\L}.~Staniszewski, K.~Zaleska, and K.~Deja, ``Low-rank continual
  personalization of diffusion models,'' \emph{arXiv preprint
  arXiv:2410.04891}, 2024.

\bibitem{flexible_customization_dm}
J.~Dong, W.~Liang, H.~Li, D.~Zhang, M.~Cao, H.~Ding, S.~H. Khan, and
  F.~Shahbaz~Khan, ``How to continually adapt text-to-image diffusion models
  for flexible customization?'' \emph{Advances in Neural Information Processing
  Systems}, vol.~37, pp. 130\,057--130\,083, 2024.

\bibitem{lora_continual_dm_deja}
{\L}.~Staniszewski, K.~Zaleska, and K.~Deja, ``Low-rank continual
  personalization of diffusion models,'' \emph{arXiv preprint
  arXiv:2410.04891}, 2024.

\bibitem{unet}
O.~Ronneberger, P.~Fischer, and T.~Brox, ``U-net: Convolutional networks for
  biomedical image segmentation,'' in \emph{Medical image computing and
  computer-assisted intervention--MICCAI 2015: 18th international conference,
  Munich, Germany, October 5-9, 2015, proceedings, part III 18}.\hskip 1em plus
  0.5em minus 0.4em\relax Springer, 2015, pp. 234--241.

\bibitem{flow_matching}
\BIBentryALTinterwordspacing
Y.~Lipman, R.~T.~Q. Chen, H.~Ben-Hamu, M.~Nickel, and M.~Le. Flow {{Matching}}
  for {{Generative Modeling}}. [Online]. Available:
  \url{http://arxiv.org/abs/2210.02747}
\BIBentrySTDinterwordspacing

\bibitem{rectified_flow}
X.~Liu, C.~Gong, and Q.~Liu, ``Flow straight and fast: Learning to generate and
  transfer data with rectified flow,'' \emph{arXiv preprint arXiv:2209.03003},
  2022.

\bibitem{stochastic_interpolant}
M.~S. Albergo and E.~Vanden-Eijnden, ``Building normalizing flows with
  stochastic interpolants,'' \emph{arXiv preprint arXiv:2209.15571}, 2022.

\bibitem{cfm_tong}
A.~Tong, K.~Fatras, N.~Malkin, G.~Huguet, Y.~Zhang, J.~Rector-Brooks, G.~Wolf,
  and Y.~Bengio, ``Improving and generalizing flow-based generative models with
  minibatch optimal transport,'' \emph{arXiv preprint arXiv:2302.00482}, 2023.

\bibitem{ddpm}
J.~Ho, A.~Jain, and P.~Abbeel, ``Denoising diffusion probabilistic models,''
  \emph{Advances in neural information processing systems}, vol.~33, pp.
  6840--6851, 2020.

\bibitem{ddim}
J.~Song, C.~Meng, and S.~Ermon, ``Denoising diffusion implicit models,''
  \emph{arXiv preprint arXiv:2010.02502}, 2020.

\bibitem{hal_continual_diffusion}
\BIBentryALTinterwordspacing
R.~Yang, M.~Grard, E.~Dellandr{\'e}a, and L.~Chen, ``{ONLINE CONTINUAL LEARNING
  OF DIFFUSION MODELS: MULTI-MODE ADAPTIVE GENERATIVE DISTILLATION},'' in
  \emph{{IEEE ICIP2025}}, Anchorage (Alaska), United States, Sep. 2025.
  [Online]. Available: \url{https://hal.science/hal-04928776}
\BIBentrySTDinterwordspacing

\bibitem{class_prototype_dm}
K.~Doan, Q.~Tran, T.~Nguyen, D.~Phung, and T.~Le, ``Class-prototype conditional
  diffusion model for continual learning with generative replay,'' \emph{arXiv
  preprint arXiv:2312.06710}, 2023.

\bibitem{cifar}
A.~Krizhevsky, G.~Hinton \emph{et~al.}, ``Learning multiple layers of features
  from tiny images.(2009),'' 2009.

\bibitem{dynamic_cl_1}
A.~Mallya and S.~Lazebnik, ``Packnet: Adding multiple tasks to a single network
  by iterative pruning,'' in \emph{Proceedings of the IEEE conference on
  Computer Vision and Pattern Recognition}, 2018, pp. 7765--7773.

\bibitem{hat}
J.~Serra, D.~Suris, M.~Miron, and A.~Karatzoglou, ``Overcoming catastrophic
  forgetting with hard attention to the task,'' in \emph{International
  conference on machine learning}.\hskip 1em plus 0.5em minus 0.4em\relax PMLR,
  2018, pp. 4548--4557.

\bibitem{den}
J.~Yoon, E.~Yang, J.~Lee, and S.~J. Hwang, ``Lifelong learning with dynamically
  expandable networks,'' \emph{arXiv preprint arXiv:1708.01547}, 2017.

\bibitem{pnn}
A.~A. Rusu, N.~C. Rabinowitz, G.~Desjardins, H.~Soyer, J.~Kirkpatrick,
  K.~Kavukcuoglu, R.~Pascanu, and R.~Hadsell, ``Progressive neural networks,''
  \emph{arXiv preprint arXiv:1606.04671}, 2016.

\bibitem{continual_learning_through_synaptic_inelligence}
F.~Zenke, B.~Poole, and S.~Ganguli, ``Continual learning through synaptic
  intelligence,'' in \emph{International conference on machine learning}.\hskip
  1em plus 0.5em minus 0.4em\relax PMLR, 2017, pp. 3987--3995.

\bibitem{ewc}
J.~Kirkpatrick, R.~Pascanu, N.~Rabinowitz, J.~Veness, G.~Desjardins, A.~A.
  Rusu, K.~Milan, J.~Quan, T.~Ramalho, A.~Grabska-Barwinska \emph{et~al.},
  ``Overcoming catastrophic forgetting in neural networks,'' \emph{Proceedings
  of the national academy of sciences}, vol. 114, no.~13, pp. 3521--3526, 2017.

\bibitem{memory_aware_synapses}
R.~Aljundi, F.~Babiloni, M.~Elhoseiny, M.~Rohrbach, and T.~Tuytelaars, ``Memory
  aware synapses: Learning what (not) to forget,'' in \emph{Proceedings of the
  European conference on computer vision (ECCV)}, 2018, pp. 139--154.

\bibitem{reg_cl_4}
S.-W. Lee, J.-H. Kim, J.~Jun, J.-W. Ha, and B.-T. Zhang, ``Overcoming
  catastrophic forgetting by incremental moment matching,'' \emph{Advances in
  neural information processing systems}, vol.~30, 2017.

\bibitem{adam_nscl}
S.~Wang, X.~Li, J.~Sun, and Z.~Xu, ``Training networks in null space of feature
  covariance for continual learning,'' in \emph{Proceedings of the IEEE/CVF
  conference on Computer Vision and Pattern Recognition}, 2021, pp. 184--193.

\bibitem{omw_orthogonal_gradient}
G.~Zeng, Y.~Chen, B.~Cui, and S.~Yu, ``Continual learning of context-dependent
  processing in neural networks,'' \emph{Nature Machine Intelligence}, vol.~1,
  no.~8, pp. 364--372, 2019.

\bibitem{orthogonal_cl_3}
G.~Saha, I.~Garg, and K.~Roy, ``Gradient projection memory for continual
  learning,'' \emph{arXiv preprint arXiv:2103.09762}, 2021.

\bibitem{orthogonal_cl_4}
H.~Sahbi and H.~Zhan, ``Ffnb: Forgetting-free neural blocks for deep continual
  visual learning,'' \emph{arXiv preprint arXiv:2111.11366}, 2021.

\bibitem{online_continual_learning}
R.~Aljundi, M.~Lin, B.~Goujaud, and Y.~Bengio, ``Gradient based sample
  selection for online continual learning,'' \emph{Advances in neural
  information processing systems}, vol.~32, 2019.

\bibitem{mem_replay_cl_2}
D.~Isele and A.~Cosgun, ``Selective experience replay for lifelong learning,''
  in \emph{Proceedings of the AAAI Conference on Artificial Intelligence},
  vol.~32, no.~1, 2018.

\bibitem{forgetting_metric}
A.~Chaudhry, P.~K. Dokania, T.~Ajanthan, and P.~H.~S. Torr, ``Riemannian walk
  for incremental learning: Understanding forgetting and intransigence,'' in
  \emph{Proceedings of the European Conference on Computer Vision (ECCV)},
  September 2018.

\bibitem{mem_replay_cl_4}
A.~Chaudhry, A.~Gordo, P.~Dokania, P.~Torr, and D.~Lopez-Paz, ``Using hindsight
  to anchor past knowledge in continual learning,'' in \emph{Proceedings of the
  AAAI conference on artificial intelligence}, vol.~35, no.~8, 2021, pp.
  6993--7001.

\bibitem{gan_cl_1}
M.~Zhai, L.~Chen, F.~Tung, J.~He, M.~Nawhal, and G.~Mori, ``Lifelong gan:
  Continual learning for conditional image generation,'' in \emph{Proceedings
  of the IEEE/CVF international conference on computer vision}, 2019, pp.
  2759--2768.

\bibitem{dgr}
H.~Shin, J.~K. Lee, J.~Kim, and J.~Kim, ``Continual learning with deep
  generative replay,'' \emph{Advances in neural information processing
  systems}, vol.~30, 2017.

\bibitem{gan}
I.~Goodfellow, J.~Pouget-Abadie, M.~Mirza, B.~Xu, D.~Warde-Farley, S.~Ozair,
  A.~Courville, and Y.~Bengio, ``Generative adversarial networks,''
  \emph{Communications of the ACM}, vol.~63, no.~11, pp. 139--144, 2020.

\bibitem{vae_cl_2}
G.~Van~de Ven and A.~Tolias, ``Generative replay with feedback connections as a
  general strategy for continual learning,'' \emph{arXiv preprint
  arXiv:1809.10635}, 2018.

\bibitem{vae}
D.~P. Kingma, ``Auto-encoding variational bayes,'' \emph{arXiv preprint
  arXiv:1312.6114}, 2013.

\bibitem{task_agnostic_cl}
P.~Kirichenko, M.~Farajtabar, D.~Rao, B.~Lakshminarayanan, N.~Levine, A.~Li,
  H.~Hu, A.~G. Wilson, and R.~Pascanu, ``Task-agnostic continual learning with
  hybrid probabilistic models,'' \emph{arXiv preprint arXiv:2106.12772}, 2021.

\bibitem{nf_kobyzev}
I.~Kobyzev, S.~J. Prince, and M.~A. Brubaker, ``Normalizing flows: An
  introduction and review of current methods,'' \emph{IEEE transactions on
  pattern analysis and machine intelligence}, vol.~43, no.~11, pp. 3964--3979,
  2020.

\bibitem{generative_feature_replay_gan}
X.~Liu, C.~Wu, M.~Menta, L.~Herranz, B.~Raducanu, A.~D. Bagdanov, S.~Jui, and
  J.~v. de~Weijer, ``Generative feature replay for class-incremental
  learning,'' in \emph{Proceedings of the IEEE/CVF Conference on Computer
  Vision and Pattern Recognition (CVPR) Workshops}, June 2020.

\bibitem{cgil}
E.~Frascaroli, A.~Panariello, P.~Buzzega, L.~Bonicelli, A.~Porrello, and
  S.~Calderara, ``Clip with generative latent replay: a strong baseline for
  incremental learning,'' \emph{arXiv preprint arXiv:2407.15793}, 2024.

\bibitem{brain_inspired_replay_vae}
G.~M. Van~de Ven, H.~T. Siegelmann, and A.~S. Tolias, ``Brain-inspired replay
  for continual learning with artificial neural networks,'' \emph{Nature
  communications}, vol.~11, no.~1, p. 4069, 2020.

\bibitem{closed_loop_transcription_incremental_vae}
S.~Tong, X.~Dai, Z.~Wu, M.~Li, B.~Yi, and Y.~Ma, ``Incremental learning of
  structured memory via closed-loop transcription,'' \emph{arXiv preprint
  arXiv:2202.05411}, 2022.

\bibitem{task_agnostic_continual_learning_nf}
P.~Kirichenko, M.~Farajtabar, D.~Rao, B.~Lakshminarayanan, N.~Levine, A.~Li,
  H.~Hu, A.~G. Wilson, and R.~Pascanu, ``Task-agnostic continual learning with
  hybrid probabilistic models,'' \emph{arXiv preprint arXiv:2106.12772}, 2021.

\bibitem{dm_replay_features}
J.~Zhang, Y.~Li, X.~Liu, and S.~Wang, ``Diffusion model meets non-exemplar
  class-incremental learning and beyond,'' \emph{arXiv preprint
  arXiv:2408.02983}, 2024.

\bibitem{coop}
K.~Zhou, J.~Yang, C.~C. Loy, and Z.~Liu, ``Learning to prompt for
  vision-language models,'' \emph{International Journal of Computer Vision},
  vol. 130, no.~9, pp. 2337--2348, 2022.

\bibitem{cocoop}
------, ``Conditional prompt learning for vision-language models,'' in
  \emph{Proceedings of the IEEE/CVF conference on computer vision and pattern
  recognition}, 2022, pp. 16\,816--16\,825.

\bibitem{coda_prompt}
J.~S. Smith, L.~Karlinsky, V.~Gutta, P.~Cascante-Bonilla, D.~Kim, A.~Arbelle,
  R.~Panda, R.~Feris, and Z.~Kira, ``Coda-prompt: Continual decomposed
  attention-based prompting for rehearsal-free continual learning,'' in
  \emph{Proceedings of the IEEE/CVF conference on computer vision and pattern
  recognition}, 2023, pp. 11\,909--11\,919.

\bibitem{dual_prompt}
Z.~Wang, Z.~Zhang, S.~Ebrahimi, R.~Sun, H.~Zhang, C.-Y. Lee, X.~Ren, G.~Su,
  V.~Perot, J.~Dy \emph{et~al.}, ``Dualprompt: Complementary prompting for
  rehearsal-free continual learning,'' in \emph{European conference on computer
  vision}.\hskip 1em plus 0.5em minus 0.4em\relax Springer, 2022, pp. 631--648.

\bibitem{attri_clip}
R.~Wang, X.~Duan, G.~Kang, J.~Liu, S.~Lin, S.~Xu, J.~L{\"u}, and B.~Zhang,
  ``Attriclip: A non-incremental learner for incremental knowledge learning,''
  in \emph{Proceedings of the IEEE/CVF Conference on Computer Vision and
  Pattern Recognition}, 2023, pp. 3654--3663.

\bibitem{star_prompt}
M.~Menabue, E.~Frascaroli, M.~Boschini, E.~Sangineto, L.~Bonicelli,
  A.~Porrello, and S.~Calderara, ``Semantic residual prompts for continual
  learning,'' in \emph{European Conference on Computer Vision}.\hskip 1em plus
  0.5em minus 0.4em\relax Springer, 2024, pp. 1--18.

\bibitem{vpt_continual}
Y.~Lu, S.~Zhang, D.~Cheng, Y.~Xing, N.~Wang, P.~Wang, and Y.~Zhang, ``Visual
  prompt tuning in null space for continual learning,'' \emph{arXiv preprint
  arXiv:2406.05658}, 2024.

\bibitem{eclipse_visual_continual}
B.~Kim, J.~Yu, and S.~J. Hwang, ``Eclipse: Efficient continual learning in
  panoptic segmentation with visual prompt tuning,'' in \emph{Proceedings of
  the IEEE/CVF Conference on Computer Vision and Pattern Recognition}, 2024,
  pp. 3346--3356.

\bibitem{clip}
A.~Radford, J.~W. Kim, C.~Hallacy, A.~Ramesh, G.~Goh, S.~Agarwal, G.~Sastry,
  A.~Askell, P.~Mishkin, J.~Clark \emph{et~al.}, ``Learning transferable visual
  models from natural language supervision,'' in \emph{International conference
  on machine learning}.\hskip 1em plus 0.5em minus 0.4em\relax PMLR, 2021, pp.
  8748--8763.

\bibitem{prompt_feature_replay_continual}
Z.~Huang, Z.~Chen, Z.~Chen, E.~Zhou, X.~Xu, R.~S.~M. Goh, Y.~Liu, W.~Zuo, and
  C.~Feng, ``Learning prompt with distribution-based feature replay for
  few-shot class-incremental learning,'' \emph{arXiv preprint
  arXiv:2401.01598}, 2024.

\bibitem{cgil_cond}
V.~Enescu and H.~Sahbi, ``Frugal incremental generative modeling using
  variational autoencoders,'' \emph{arXiv preprint arXiv:2505.22408}, 2025.

\bibitem{slca}
G.~Zhang, L.~Wang, G.~Kang, L.~Chen, and Y.~Wei, ``Slca: Slow learner with
  classifier alignment for continual learning on a pre-trained model,'' in
  \emph{Proceedings of the IEEE/CVF International Conference on Computer
  Vision}, 2023, pp. 19\,148--19\,158.

\bibitem{slca++}
------, ``Slca++: Unleash the power of sequential fine-tuning for continual
  learning with pre-training,'' \emph{arXiv preprint arXiv:2408.08295}, 2024.

\bibitem{pilora}
H.~Guo, F.~Zhu, W.~Liu, X.-Y. Zhang, and C.-L. Liu, ``Pilora: Prototype guided
  incremental lora for federated class-incremental learning,'' in
  \emph{European Conference on Computer Vision}.\hskip 1em plus 0.5em minus
  0.4em\relax Springer, 2024, pp. 141--159.

\bibitem{sd_lora}
Y.~Wu, H.~Piao, L.-K. Huang, R.~Wang, W.~Li, H.~Pfister, D.~Meng, K.~Ma, and
  Y.~Wei, ``S-lora: Scalable low-rank adaptation for class incremental
  learning,'' \emph{arXiv preprint arXiv:2501.13198}, 2025.

\bibitem{o_lora}
\BIBentryALTinterwordspacing
X.~Wang, T.~Chen, Q.~Ge, H.~Xia, R.~Bao, R.~Zheng, Q.~Zhang, T.~Gui, and
  X.~Huang, ``Orthogonal subspace learning for language model continual
  learning,'' in \emph{Findings of the Association for Computational
  Linguistics: EMNLP 2023}, H.~Bouamor, J.~Pino, and K.~Bali, Eds.\hskip 1em
  plus 0.5em minus 0.4em\relax Singapore: Association for Computational
  Linguistics, Dec. 2023, pp. 10\,658--10\,671. [Online]. Available:
  \url{https://aclanthology.org/2023.findings-emnlp.715/}
\BIBentrySTDinterwordspacing

\bibitem{c_lora}
X.~Zhang, L.~Bai, X.~Yang, and J.~Liang, ``C-lora: Continual low-rank
  adaptation for pre-trained models,'' \emph{arXiv preprint arXiv:2502.17920},
  2025.

\bibitem{lori}
J.~Zhang, J.~You, A.~Panda, and T.~Goldstein, ``Lori: Reducing cross-task
  interference in multi-task low-rank adaptation,'' \emph{arXiv preprint
  arXiv:2504.07448}, 2025.

\bibitem{critical_param_lora}
S.~Ling, L.~Zhang, J.~Zhao, L.~Pan, and H.~Li, ``Lora-based continual learning
  with constraints on critical parameter changes,'' \emph{arXiv preprint
  arXiv:2504.13407}, 2025.

\bibitem{fm_lora}
X.~Yu, J.~Yang, X.~Wu, P.~Qiu, and X.~Liu, ``Fm-lora: Factorized low-rank
  meta-prompting for continual learning,'' \emph{arXiv preprint
  arXiv:2504.08823}, 2025.

\bibitem{dc_lora}
\BIBentryALTinterwordspacing
L.~Li, S.~Wang, C.~Li, Y.~Yuan, and G.~Wang, ``Dc-lora: Domain correlation
  low-rank adaptation for domain incremental learning,'' \emph{High-Confidence
  Computing}, p. 100270, 2025. [Online]. Available:
  \url{https://www.sciencedirect.com/science/article/pii/S2667295224000734}
\BIBentrySTDinterwordspacing

\bibitem{orthogonal_lora_lie_cl}
K.~Cao and S.~Wu, ``Orthogonal low-rank adaptation in lie groups for continual
  learning of large language models,'' \emph{arXiv preprint arXiv:2509.06100},
  2025.

\bibitem{lora_critical_parameter_change_cl}
S.~Ling, L.~Zhang, J.~Zhao, L.~Pan, and H.~Li, ``Lora-based continual learning
  with constraints on critical parameter changes,'' \emph{arXiv preprint
  arXiv:2504.13407}, 2025.

\bibitem{tree_lora_cl}
Y.-Y. Qian, Y.-Z. Xu, Z.-Y. Zhang, P.~Zhao, and Z.-H. Zhou, ``Treelora:
  Efficient continual learning via layer-wise loras guided by a hierarchical
  gradient-similarity tree,'' \emph{arXiv preprint arXiv:2506.10355}, 2025.

\bibitem{inflora}
Y.-S. Liang and W.-J. Li, ``Inflora: Interference-free low-rank adaptation for
  continual learning,'' in \emph{Proceedings of the IEEE/CVF Conference on
  Computer Vision and Pattern Recognition}, 2024, pp. 23\,638--23\,647.

\bibitem{gan_memory}
Y.~Cong, M.~Zhao, J.~Li, S.~Wang, and L.~Carin, ``Gan memory with no
  forgetting,'' \emph{Advances in neural information processing systems},
  vol.~33, pp. 16\,481--16\,494, 2020.

\bibitem{lfs_gan}
J.~Seo, J.-S. Kang, and G.-M. Park, ``Lfs-gan: Lifelong few-shot image
  generation,'' in \emph{Proceedings of the IEEE/CVF international conference
  on computer vision}, 2023, pp. 11\,356--11\,366.

\bibitem{kernel_modulation}
Y.~Zhao, K.~Chandrasegaran, M.~Abdollahzadeh, and N.-M.~M. Cheung, ``Few-shot
  image generation via adaptation-aware kernel modulation,'' \emph{Advances in
  Neural Information Processing Systems}, vol.~35, pp. 19\,427--19\,440, 2022.

\bibitem{adam_few_shot}
Y.~Zhao, K.~Chandrasegaran, M.~Abdollahzadeh, C.~Du, T.~Pang, R.~Li, H.~Ding,
  and N.-M. Cheung, ``Adam: Few-shot image generation via adaptation-aware
  kernel modulation,'' \emph{arXiv preprint arXiv:2307.01465}, 2023.

\bibitem{node_nf}
R.~T. Chen, Y.~Rubanova, J.~Bettencourt, and D.~K. Duvenaud, ``Neural ordinary
  differential equations,'' \emph{Advances in neural information processing
  systems}, vol.~31, 2018.

\bibitem{ot_map}
R.~J. McCann, ``A convexity principle for interacting gases,'' \emph{Advances
  in mathematics}, vol. 128, no.~1, pp. 153--179, 1997.

\bibitem{imagenet}
J.~Deng, W.~Dong, R.~Socher, L.-J. Li, K.~Li, and L.~Fei-Fei, ``Imagenet: A
  large-scale hierarchical image database,'' in \emph{2009 IEEE conference on
  computer vision and pattern recognition}.\hskip 1em plus 0.5em minus
  0.4em\relax Ieee, 2009, pp. 248--255.

\bibitem{cil_survey_table}
L.~Wang, X.~Zhang, H.~Su, and J.~Zhu, ``A comprehensive survey of continual
  learning: Theory, method and application,'' \emph{IEEE Transactions on
  Pattern Analysis and Machine Intelligence}, 2024.

\bibitem{one_cycle_lr}
L.~N. Smith and N.~Topin, ``Super-convergence: Very fast training of neural
  networks using large learning rates,'' in \emph{Artificial intelligence and
  machine learning for multi-domain operations applications}, vol. 11006.\hskip
  1em plus 0.5em minus 0.4em\relax SPIE, 2019, pp. 369--386.

\bibitem{scaling_and_shifting}
D.~Lian, D.~Zhou, J.~Feng, and X.~Wang, ``Scaling \& shifting your features: A
  new baseline for efficient model tuning,'' \emph{Advances in Neural
  Information Processing Systems}, vol.~35, pp. 109--123, 2022.

\bibitem{bitfit}
E.~B. Zaken, S.~Ravfogel, and Y.~Goldberg, ``Bitfit: Simple parameter-efficient
  fine-tuning for transformer-based masked language-models,'' \emph{arXiv
  preprint arXiv:2106.10199}, 2021.

\bibitem{peft_normalization}
W.~Qi, Y.-P. Ruan, Y.~Zuo, and T.~Li, ``Parameter-efficient tuning on layer
  normalization for pre-trained language models,'' \emph{arXiv preprint
  arXiv:2211.08682}, 2022.

\bibitem{cocodataset}
\BIBentryALTinterwordspacing
T.~Lin, M.~Maire, S.~J. Belongie, L.~D. Bourdev, R.~B. Girshick, J.~Hays,
  P.~Perona, D.~Ramanan, P.~Doll{'{a} }r, and C.~L. Zitnick, ``Microsoft
  {COCO:} common objects in context,'' \emph{CoRR}, vol. abs/1405.0312, 2014.
  [Online]. Available: \url{http://arxiv.org/abs/1405.0312}
\BIBentrySTDinterwordspacing

\bibitem{fid_github}
\BIBentryALTinterwordspacing
A.~Obukhov, M.~Seitzer, P.-W. Wu, S.~Zhydenko, J.~Kyl, and E.~Y.-J. Lin,
  ``High-fidelity performance metrics for generative models in pytorch,'' 2020,
  version: 0.3.0, DOI: 10.5281/zenodo.4957738. [Online]. Available:
  \url{https://github.com/toshas/torch-fidelity}
\BIBentrySTDinterwordspacing

\bibitem{FID}
M.~Heusel, H.~Ramsauer, T.~Unterthiner, B.~Nessler, and S.~Hochreiter, ``Gans
  trained by a two time-scale update rule converge to a local nash
  equilibrium,'' \emph{Advances in neural information processing systems},
  vol.~30, 2017.

\end{thebibliography}

	\onecolumn
	
	\clearpage

	\section*{Supplementary Material}

	\subsection{Visualizations}
	
\begin{figure}[H]
		\centering
		\begin{subfigure}{0.45\columnwidth}
			\centering
			\includegraphics[width=1.0\columnwidth]{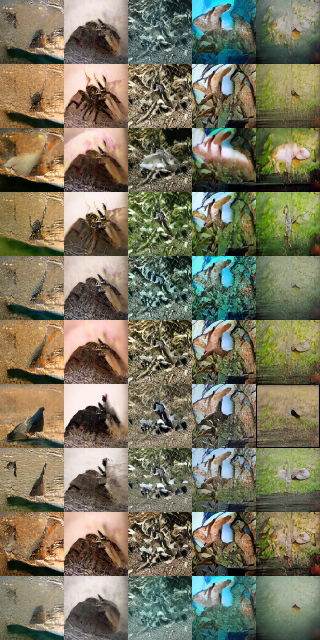}
			\caption{FunLoRA $\Fc_{\y, 10}^{\mycosfreq}$ (incremental).}
			\label{fig:incremental_samples_funlora_imagenet100}
		\end{subfigure} \quad
		\begin{subfigure}{0.45\columnwidth}
			\centering
			\includegraphics[width=1.0\columnwidth]{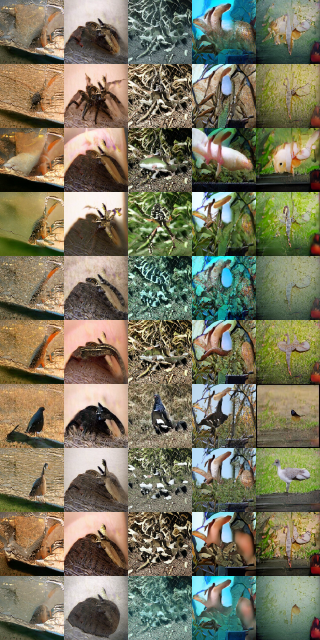}
			\caption{Multi-task (not incremental).}
			\label{fig:not_incremental_samples_funlora_imagenet100}
		\end{subfigure}
		\caption{Comparison between samples obtained using an incremental flow-matching model in Figure~\ref{fig:incremental_samples_funlora_imagenet100}, and a non-incremental one in Figure~\ref{fig:not_incremental_samples_funlora_imagenet100}, on the ImageNet100 dataset, split into 5 equal tasks. Samples obtained using the proposed  $\Fc_{\y, 10}^{\mycosfreq}$ are almost identical (visually) to the generative model trained in multi-task setting, which confirms the strength of the proposed method. The last accuracy in the multi-task and incremental settings are respectively 62.95\% and 58.30\%, and the extra parameters per class for $\Fc_{\y, 10}^{\mycosfreq}$ is equal to 13.65K. The total number of parameters of the generative model after training on the last task is 90.69M for $\Fc_{\y, 10}^{\mycosfreq}$ vs 89.59M for the multitask model.}
		\label{fig:samples_funlora_imagenet100}
	\end{figure}

\begin{figure}[H]
		\centering
		\begin{subfigure}{0.47\columnwidth}
			\centering
			\includegraphics[width=1.0\columnwidth]{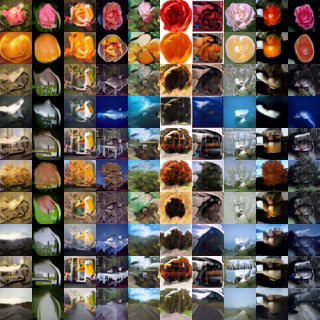}
			\caption{FunLoRA $\Fc_{\y, 10}^{\mycosfreq}$ (incremental).}
			\label{fig:incremental_samples_funlora_cifar100}
		\end{subfigure} \quad
		\begin{subfigure}{0.47\columnwidth}
			\centering
			\includegraphics[width=1.0\columnwidth]{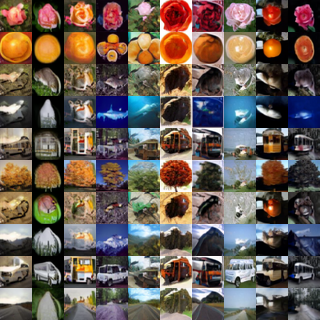}
			\caption{Multi-task (not incremental).}
			\label{fig:not_incremental_samples_funlora_cifar100}
		\end{subfigure}
		\caption{Comparison between samples obtained using an incremental flow-matching model in Figure~\ref{fig:incremental_samples_funlora_cifar100}, and a non-incremental one in Figure~\ref{fig:not_incremental_samples_funlora_cifar100}, on the CIFAR100 dataset split into 5 equal tasks. Samples obtained using the proposed  $\Fc_{\y, 10}^{\mycosfreq}$ are almost identical (visually) to the generative model trained in multi-task setting, which confirms the strength of the proposed method. The last accuracy in the multi-task and incremental settings are respectively 64.51\% and 60.07\%, and the extra parameters per class for $\Fc_{\y, 10}^{\mycosfreq}$ is equal to 10.03K. The total number of parameters of the generative model after training on the last task is 36.61M for $\Fc_{\y, 10}^{\mycosfreq}$ vs 35.81M for the multitask model.}
		\label{fig:samples_funlora_cifar100}
	\end{figure}

	\noindent \textbf{Visualizations.} In  Figure~\ref{fig:samples_funlora_imagenet100} and ~\ref{fig:samples_funlora_cifar100}, we compare images obtained using an incremental generative model trained using $\Fc_{\y, 10}^{\mycosfreq}$, with images obtained using a multi-task flow-matching model. It can be noted that the images are almost identical visually speaking, which confirms the expressiveness of the proposed methods. Furthermore, the extra number of parameters  required to learn the new classes (80 of them) is negligible in both cases, and corresponds to $1.2\%$ and $2.2\%$ of the vanilla model parameters on respectively ImageNet100 and CIFAR100. 
	

	\subsection{Training and sampling time}
	
	\begin{table}[H]
		\centering
		\resizebox{0.75\linewidth}{!}{%
			\begin{tabular}{lcc}
				\toprule
				& \begin{tabular}[c]{@{}c@{}}CIFAR10-100\end{tabular} 
				& \begin{tabular}[c]{@{}c@{}}ImageNet100\end{tabular}  \\ 
				\midrule
				Training time of the generative model (all tasks) & 3.5h & 31h \\ 
				\midrule
				Sampling time (all tasks) & 1.4h & 16.8h  \\
				\midrule
				CNN training time (all tasks) & 0.2h & 1.6h \\
				\bottomrule
			\end{tabular}
		}
		\caption{Training time, sampling time as well as CNN training time for all tasks (cumulated time) on CIFAR10 with split 2-2, CIFAR100 with split 20-20, and ImagetNet100 with split 20-20, with the default dopri5 solver. Results are reported on a NVIDIA A100 GPU with 80GB of RAM. For information, the results on ImageNet100 were run on a H100 GPU 80GB, and then multiplied by 2 to get the equivalent on a A100 GPU.}
		\label{table:sampling_time}
	\end{table}

\begin{table}[H]
	\centering
	\resizebox{0.7\linewidth}{!}{%
		\begin{tabular}{lccc} 
			\toprule
			Method        & Time [GPU-hours] $\downarrow$ & LA $\uparrow$    & Parameters in Millions $\downarrow$ \\ 
			\midrule
			DDGR~\cite{ddgr}          & 20.6h            & 43.69   & 52.4M \\
			DGR diffusion~\cite{dgr} & 14.0h            & 59.00  & 52.4M \\
			GUIDE~\cite{guide_diffusion}         & 15.30            & 64.47  & 52.4M \\
			JDCL (fast)~\cite{jdcl}   & 13.6h            & 78.10  & 52.4M \\
			JDCL~\cite{jdcl}          & 26.9h           & 83.69  & 52.4M \\
			FunLoRA (ours) & \textbf{5.1h} & \textbf{84.02} & \textbf{35.84M} \\
			\bottomrule
		\end{tabular}
\	}
	\caption{Runtime analysis of our method against closely related works on CIFAR10 with 2-2 split. It can be noted that our method achieves better accuracy scores, while requiring less overall compute and parameters. The results from this table are taken from~\cite{jdcl}, and the time [GPU-hours] is divided by 2, to get an equivalent on A100 GPU with 80GB for fair comparison (indeed, results were initially reported on A100 GPU with 40GB).}
	\label{tab:runtime_analysis}
\end{table}

	\pagebreak
	\clearpage
	
	\subsection{Image quality analysis}	
	\begin{table}[H]
		\centering
		\resizebox{0.7\linewidth}{!}{%
				\begin{tabular}{ccccccc} 
				\toprule
				\multirow{2}{*}{\textbf{Method}} & \multicolumn{3}{c}{\begin{tabular}[c]{@{}c@{}}\textbf{CIFAR10}\\\textbf{2-2}\end{tabular}} & \multicolumn{3}{c}{\begin{tabular}[c]{@{}c@{}}\textbf{CIFAR-100}\\\textbf{20-20}\end{tabular}}  \\ 
				\cmidrule(l){2-7}
				& FID $\downarrow$            & Precision $\uparrow$     & Recall  $\uparrow$                                                 & FID $\downarrow$            & Precision  $\uparrow$   & Recall     $\uparrow$                                                  \\ 
				\midrule
				DGR~\cite{dgr}                   & 49.56          & 0.54          & 0.34                                                      & 50.36         & 0.48          & 0.30                                                            \\
				JDCL~\cite{jdcl}                 & 32.12          & 0.60          & \textbf{0.43}                                             & 39.93         & \textbf{0.71} & 0.27                                                            \\
				FunLoRA (ours)                   & \textbf{14.17} & \textbf{0.67} & 0.41                                                      & \textbf{8.64} & \textbf{0.71} & \textbf{0.45}                                                   \\
				\bottomrule
			\end{tabular}
		}
		\caption{Comparison of the image quality obtained using FunLoRA with~\cite{jdcl} and~\cite{ddgr}, on CIFAR10 with split 2-2, and CIFAR100 with split 20-20. Our results in this table are obtained using~\cite{fid_github} github, and the other are taken from~\cite{jdcl}. It can be noted the the FID~\cite{FID} scores obtained using FunLoRA are consistently better than other methods at the end of the incremental training.}
		\label{tab:generativ_cl}
	\end{table}
	
	\subsection{Extra results with pre-training on COCO dataset} \label{results_with_pretraining_on_coco}
	
	\begin{table*}[!h]
		\centering
		\resizebox{\linewidth}{!}{%
			\begin{tabular}{lcccccccc} 
				\toprule
				& CIFAR10                  & \multirow{2}{*}{P\#} & \multicolumn{2}{c}{CIFAR100}                        & \multirow{2}{*}{P\#} & ImageNet100              & \multirow{2}{*}{P\#} & \multirow{2}{*}{\begin{tabular}[c]{@{}c@{}}Pre-trained\\SD\end{tabular}}  \\ 
				\cmidrule(lr){2-2}\cmidrule(l){4-5}\cmidrule(lr){7-7}
				Dataset Split                                                        & 2-2                      &                      & 20-20                    & 10-10                    &                      & 20-20                    &                      &                                                                          \\ 
				\midrule
				DDGR \cite{ddgr}                                                     & $\idf{43.69_{\pm 2.60}}$ & 52.4M                & $\idf{28.11_{\pm 2.58}}$ & $\idf{15.99_{\pm 1.08}}$ & 52.4M                & $\idf{25.59_{\pm 2.29}}$ & 295M                 & \xmark                                                                   \\
				DGR Diffusion \cite{dgr}                                             & $\idf{59.00_{\pm 0.57}}$ & 52.4M                & $\idf{28.25_{\pm 0.22}}$ & $\idf{15.90_{\pm 1.01}}$ & 52.4M                & $\idf{23.92_{\pm 0.92}}$ & 295M                 & \xmark                                                                   \\
				GUIDE \cite{guide_diffusion}                               & $\idf{64.47_{\pm 0.45}}$ & 52.4M                & $\idf{41.66_{\pm 0.40}}$ & $\idf{26.13_{\pm 0.29}}$ & 52.4M                & $\idf{39.07_{\pm 1.37}}$ & 295M                 & \xmark                                                                   \\
				JDCL \cite{jdcl}                                                     & $\idf{83.69_{\pm 1.44}}$ & 52.4M                & $\idf{47.05_{\pm 0.61}}$ & $\idf{29.04_{\pm 0.41}}$ & 52.4M                & $\idf{54.53_{\pm 2.15}}$ & 295M                 & \xmark                                                                   \\
				DiffClass \cite{class_incremental_diffusion_pretraind_gaussian}      & NA                       & 983M                 & $\idf{62.21}$            & $\idf{58.40}$            & 983M                 & $\idf{67.26}$            & 983M                 & \cmark                                                                   \\ 
				\midrule
				FunLoRA: $\Fc_{\y, 10}^{\mycosfreq}$ (ours)                                                       & $\idf{84.02_{\pm 0.44}}$ & \textbf{35.84M}               & 
				$\idf{60.07_{\pm 0.87}}$ & $\idf{57.46_{\pm 0.55}}$                        & \textbf{36.61M}                & $\idf{58.30_{\pm 0.6}}$ & \textbf{90.69M}               & \xmark                                                                   \\
				\midrule
				\begin{tabular}[c]{@{}l@{}}FunLoRA:  $\Fc_{\y, 10}^{\mycosfreq}$ + \\ Resample (ours)\end{tabular} & $\idf{87.71_{\pm 0.36}}$                        & \textbf{35.84M}               & $\idf{67.89_{\pm 0.29}}$                        & $\idf{64.55_{\pm 0.6}}$                        & \textbf{36.61M}               & $\idf{63.03 _{\pm 0.69}}$ & \textbf{90.69M}              & \xmark                                                                   \\
				\midrule
				\begin{tabular}[c]{@{}l@{}}FunLoRA:  $\Fc_{\y, 10}^{\mycosfreq}$ + \\ Pre-train  (ours)\end{tabular} & $\idf{86.3_{\pm 0.15}}$                        & \textbf{35.84M}               & $\idf{63.19_{\pm 0.23}}$                        & $\idf{62.7_{\pm 0.24}}$                        & \textbf{36.61M}               & $\idf{62.72_{\pm 0.52}
				}$ & \textbf{90.69M}              & \xmark                                                                   \\
				\midrule
				\begin{tabular}[c]{@{}l@{}}FunLoRA:  $\Fc_{\y, 10}^{\mycosfreq}$ + \\ Pre-train  \\ + Resample (ours)\end{tabular} & $\mathbf{\idf{90.05_{\pm 0.1}}}$                        & \textbf{35.84M}               & $\mathbf{\idf{72.22_{\pm 0.42}}}$                        & $\mathbf{\idf{70.01_{\pm 1.77}}}$                        & \textbf{36.61M}               & $\mathbf{\idf{67.39_{\pm 0.18}}}$ & \textbf{90.69M}              & \xmark                                                                   \\
				\bottomrule
			\end{tabular}
		}
		\caption{Same results as in Table~\ref{table:final_results_table_lora_avg_incr_acc}, but also with 2 extra set of scores pre-training the U-Net backbone on COCO~\cite{cocodataset} dataset with unlabelled data (unsupervised training). This pre-training, greatly increases results, and the best results are also achieved on ImageNet100 as well — outperforming \cite{class_incremental_diffusion_pretraind_gaussian} using a pre-trained stable diffusion model.}
		\label{table:final_results_table_lora_avg_incr_acc_pretrain}
	\end{table*}

\begin{table}
	\centering
	\resizebox{0.9\linewidth}{!}{%
		\begin{tabular}{lcccccc} 
			\toprule
			\multirow{2}{*}{Dataset}                                                                                                      & CIFAR10~        & \multirow{2}{*}{PPC} & CIFAR100         & \multirow{2}{*}{PPC} & ImageNet100~     & \multirow{2}{*}{PPC}  \\ 
			\cmidrule(lr){2-2}\cmidrule(lr){4-4}\cmidrule(lr){6-6}
			& 2-2             &                      & 20-20            &                      & 20-20            &                       \\ 
			\midrule
			FunLoRA $+ \Fc_{\y, 10}^{\mycosfreq}$                                                                                         & $\mathbf{84.02_{\pm 0.44}}$ &  10.03K               & $\mathbf{60.07_{\pm 0.87}}$ &  10.03K               & $\mathbf{58.30_{\pm 0.6}}$ & 13.65K                  \\ 
			\midrule
			\begin{tabular}[c]{@{}l@{}}FunLoRA $+ \Fc_{\y, 10}^{\mycosfreq}$~\\+ pre-train COCO\\+ square root factorization\end{tabular} & $83.43_\pm 0.4$ & \textbf{2.474K}               & $58.92_\pm 0.54$ & \textbf{2.474K}               & $57.81_\pm 0.19$ & \textbf{3.6K}                  \\
			\bottomrule
		\end{tabular}
	}
	\caption{In this Table, we investigate a square root factorization for the LoRA parameters, so that we need only one reparametrized matrix for all layers. Indeed, if we denote $n$ as the total number of parameters to be adapted with LoRA, we define $\Ab_{\y}$ and $\Bb_{\y}$ with a dimension of $\sqrt{n}$ (rounded to the superior value). Then, we only have to increase the rank of the matrix $F_{\y}  = \Ab_{\y} \Bb_{\y}$ matrix using FunLoRA, since the indices for each layer can simply be indexed afterwards. This method allows to save a lot of parameters, and obtains results very similar to the ones in table~\ref{table:closest_comparative_lora_methods}, if we allow a COCO pre-training for the U-Net backbone. Indeed, it barely looses 1 accuracy point while using up to 4 times less parameters per class (PPC).}
	\label{table_sqr_factorization}
\end{table}
	
	\raggedbottom
	
	\pagebreak
	\clearpage
	
	\subsection{Further analysis of reparametrized matrix rank}
	
	\begin{figure*}[!b]
		\includegraphics[width=1.0\textwidth]{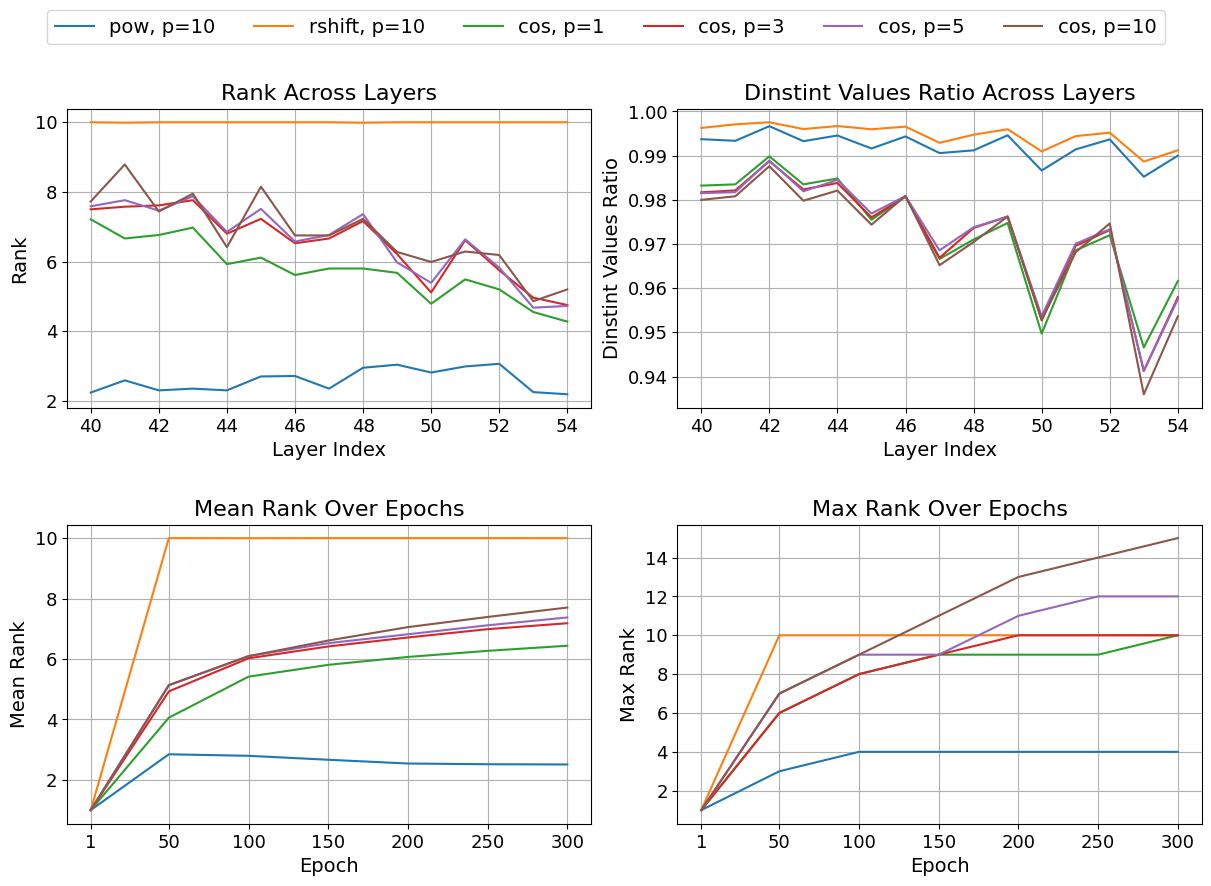}
		\centering
		\caption{i)This table provides a comprehensive analysis of LoRA matrix characteristics: (i) the average rank of the LoRA matrix across all layers (indices 40 to 54) is presented in the upper left; (ii) the ratio between the number of distinct values in each LoRA matrix and the total number of values is shown in the upper right; (iii) the average LoRA matrix rank across all layers as the number of epochs increases is depicted in the lower left; and (iv) the maximum LoRA matrix rank across all layers as the number of epochs increases is in the lower right. Notably, the "rshift-10" operation consistently increases the rank to 10, corresponding to the total number of distinct functions applied, and yields the most significant number of distinct values in each LoRA matrix. Conversely, the "cosine" operation achieves the highest maximum rank, potentially allowing for greater localized expressiveness, whereas the "power" function only reaches a maximum rank of 4.}
		\label{fig:combined_plots_funlora_full}
	\end{figure*}
	
	\raggedbottom
	
\end{document}